\pdfoutput=1

\documentclass[11pt]{article}

\usepackage[final]{acl}

\usepackage{times}
\usepackage{latexsym}

\usepackage[T1]{fontenc}

\usepackage[utf8]{inputenc}

\usepackage{microtype}

\usepackage{inconsolata}

\usepackage{graphicx}
\usepackage{amsmath}
\usepackage{amssymb}
\usepackage{mathtools}
\usepackage{amsthm}
\usepackage{hyperref}
\usepackage{amssymb}
\usepackage{multirow}
\usepackage{xcolor}
\usepackage{colortbl}
\usepackage{booktabs}

\usepackage{algorithm}
\usepackage{algorithmicx}
\usepackage{algpseudocode}
\usepackage{verbatim}
\usepackage{arydshln}
\usepackage{caption}
\usepackage{subcaption}

\definecolor{RED}{HTML}{E58579}
\definecolor{GREEN}{HTML}{9DD0C7}

\title{\centering Analyzing the Effects of Supervised Fine-Tuning on Model Knowledge\\from Token and Parameter Levels}

\author{
\bf{
Junjie Ye$^{1}$\thanks{Equal Contribution.}, Yuming Yang$^{1\ast}$, Yang Nan$^1$, Shuo Li$^1$, Qi Zhang$^{1,3}$,}\\
\bf{Tao Gui$^{1,3,4}$\thanks{Corresponding Author.}, Xuanjing Huang$^{1,3}$, Peng Wang$^{2}$, Zhongchao Shi$^{2}$, Jianping Fan$^{2}$}\vspace{2mm}\\
{$^1$Fudan University}\ \ 
{$^2$Lenovo Research, Beijing, China}\\
{$^3$Shanghai Key Lab of Intelligent Information Processing}\\
{$^4$Shanghai Innovation Institute}\vspace{2mm}\\
\texttt{jjye23@m.fudan.edu.cn, tgui@fudan.edu.cn} 
}

\begin{document}
\maketitle
\begin{abstract}

Large language models (LLMs) acquire substantial world knowledge during pre-training, which is further shaped by post-training techniques such as supervised fine-tuning (SFT). However, the impact of SFT on a model's knowledge remains underexplored, limiting our ability to control knowledge change behavior in fine-tuned models. To address this gap, we evaluate closed-book question answering (CBQA) performance across five LLMs from the LLaMA-2 and LLaMA-3 families. Surprisingly, models fine-tuned on 1,920 samples perform up to 14\% worse than those fine-tuned on only 240 samples. Furthermore, varying the level of knowledge mastery in the fine-tuning data leads to performance fluctuations of over 12\%. To investigate these effects, we analyze model behavior at both the token and parameter levels. Our analysis reveals that up to 90\% of parameter updates during SFT do not contribute to knowledge enhancement. Restoring these updates can improve performance on the CBQA task, depending on the characteristics of the fine-tuning data. These insights offer practical guidance for developing fine-tuning strategies that more effectively strengthen model knowledge.
\end{abstract}

\section{Introduction}

Large language models (LLMs)~\cite{Claude, GPT-4, LLaMA3.1, Qwen2.5} acquire extensive world knowledge through pre-training on massive text corpora~\cite{analysis-chen, analy-ye}. This knowledge is subsequently shaped through post-training techniques such as supervised fine-tuning (SFT)~\cite{SFT-1} and reinforcement learning~\cite{rlhf}, enabling LLMs to perform diverse downstream tasks, including reading comprehension~\cite{LLMs-MRC-1}, code generation~\cite{CodeLLaMA},  and tool use~\cite{RoTBench, TL-Training}.

Recent research has explored how model knowledge evolves during training. For instance, pre-training has been shown to encode knowledge modularly~\cite{knowledge}, with each parameter storing up to 2 bits of information~\cite{Physics}. Conversely, instruction fine-tuning may increase hallucinations~\cite{sft-google, SFT-2}. Empirical evidence suggests that preserving the distribution of internal representations is crucial to maintaining performance~\cite{sft-renmin}, and models with richer knowledge can be easier to fine-tune for enhanced reasoning ability~\cite{LIMO}.

\begin{figure}[!t]
    \centering
    \includegraphics[width=\linewidth]{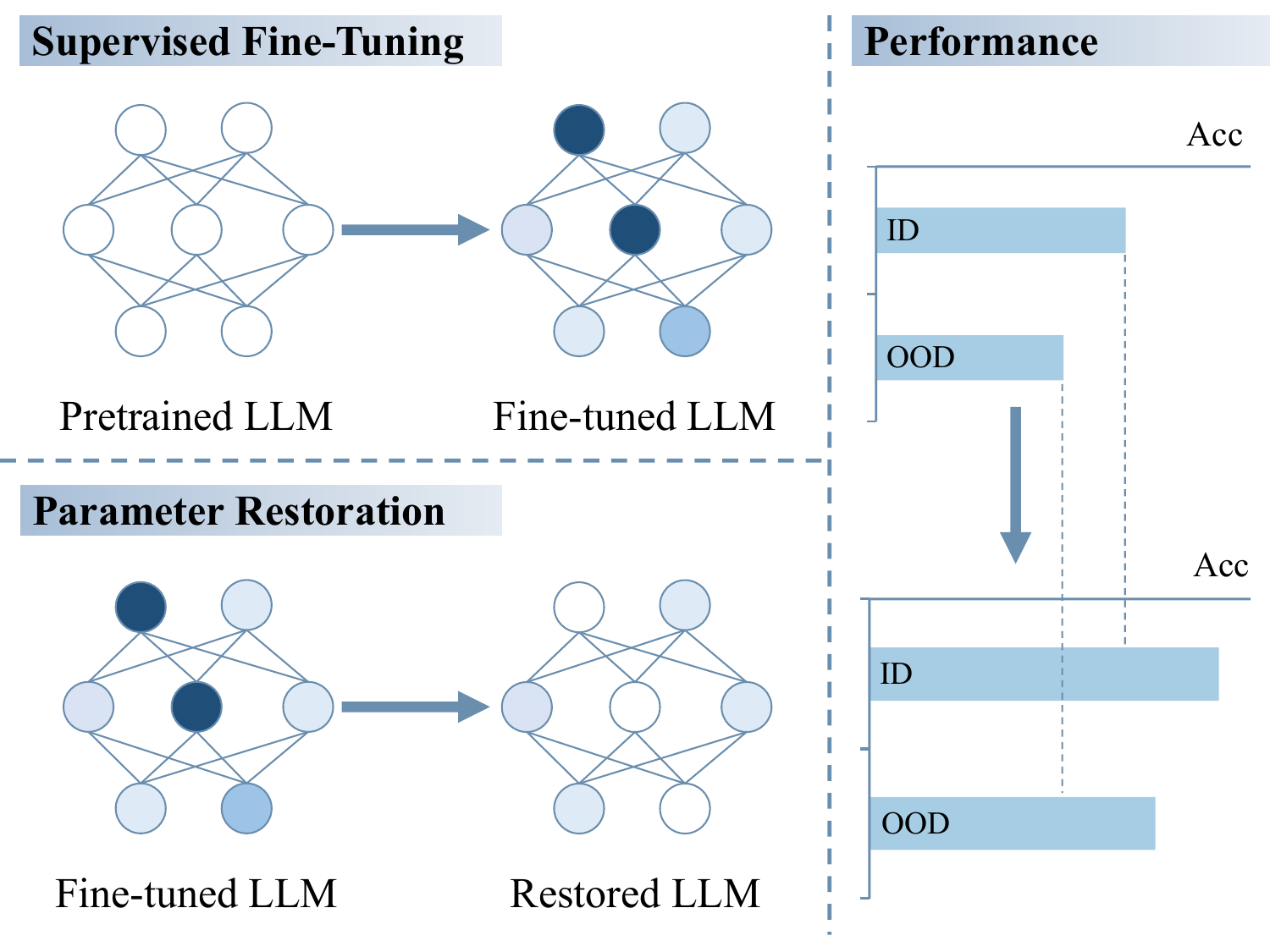}
    \caption{Illustration of parameter restoration. We find that SFT introduces many unnecessary parameter updates, and model performance can be significantly improved by restoring some of the most updated parameters in the fine-tuned model to their original values in the pre-trained model.}
    \label{fig:illustration-restore}
\end{figure}

Despite these insights, the specific impact of SFT on model knowledge remains insufficiently understood. Key open questions include how model knowledge changes with different categories and scales of fine-tuning data, the mechanisms behind these changes, and strategies to mitigate undesirable effects. This gap limits our ability to predict and control knowledge change behavior in fine-tuned models.

To address this, we evaluate five LLMs from the LLaMA-2 and LLaMA-3 families on the closed-book question answering (CBQA) task. We categorize fine-tuning data into five groups based on the knowledge mastery level and systematically examine how performance varies across these categories and different data scales. Surprisingly, models fine-tuned with 1,920 samples perform up to 14\% worse than those fine-tuned with only 240 samples. Moreover, performance fluctuates by over 12\% depending on the data category used.

To investigate these discrepancies, we conduct a token-level analysis by computing the Kullback-Leibler (KL) divergence~\cite{KL} between token logits of fine-tuned and pre-trained models (Section~\ref{sec:token-level}).
Our results show that as fine-tuning data size increases, KL divergence initially decreases, reflecting reduced deviation from the pre-trained model. However, beyond a threshold, KL divergence sharply rises, especially when fine-tuning on poorly mastered data, correlating with performance degradation.

Building on these findings, we perform a parameter-level analysis (Section~\ref{sec:parameter-level}) by selectively restoring parameters that changed most during SFT back to their pre-trained values (Figure~\ref{fig:illustration-restore}). We observe that restoring up to 90\% of parameter updates does not harm and can even improve performance on training and test sets, with improvements exceeding 10\% in some cases. This indicates that many SFT-induced updates are unnecessary for knowledge enhancement, suggesting new directions for optimizing fine-tuning.

In summary, our contributions are as follows:
\begin{itemize}
    \item We conduct extensive experiments on the CBQA task and reveal surprising effects of fine-tuning data category and scale on model knowledge.
\item Through token-level and parameter-level analyses, we find that 90\% of the parameter updates from fine-tuning do not contribute to knowledge enhancement.
\item We demonstrate that restoring these parameters improves model performance, offering practical guidance for more effective fine-tuning strategies.
\end{itemize}

\section{Related Work}

\paragraph{CBQA and Model Knowledge}
The CBQA task evaluates an LLM's ability to answer factual questions using its internal knowledge, without relying on external reference materials~\cite{QA-1, QA-2, QA-3}. This makes CBQA a rigorous test of the model's knowledge accuracy and completeness.
One significant challenge in CBQA is addressing hallucinations-instances where the model generates incorrect or fabricated answers~\cite{Hallucination, Kandpal2022LargeLM, emnlp/KangC23}. To mitigate hallucinations and enhance performance, several strategies have been proposed. For instance,~\citet{sft-renmin} investigate the impact of fine-tuning on the consistency of a model's pre-existing knowledge, emphasizing the need for stable knowledge retention during fine-tuning. Similarly,~\citet{sft-google} identify overfitting to fine-tuning data as a major source of hallucinations, noting that fine-tuning with data unfamiliar to the model exacerbates this issue. Additionally,~\citet{sft-ye} examine how variations in dataset size and quality influence CBQA outcomes, highlighting the trade-offs between data volume and model performance.
Despite these advances, prior studies primarily focus on dataset characteristics and overlook the fine-tuning process's internal dynamics. In contrast, our work provides a detailed analysis at both the token and parameter levels, identifying unnecessary parameter updates during fine-tuning as a key factor contributing to performance degradation on CBQA. 

\paragraph{Data Quality and Scale of SFT}
SFT plays a pivotal role in adapting LLMs to labeled data, enabling strong performance on downstream tasks. Consequently, constructing high-quality fine-tuning datasets is critical for maximizing SFT's effectiveness~\cite{sft-quality-1, sft-quality-2, sft-quality-3}.
Recent research highlights the effectiveness of SFT with small, high-quality datasets, achieving performance on par with larger datasets~\cite{LIMA, Beyond}. High-quality data is typically characterized as accurate, diverse, and complex~\cite{sft-quality-4, sft-quality-5, sft-quality-6, Yangdiver}, prompting efforts to synthesize such datasets automatically~\cite{Baize, Magpie, MiniGPT-4}. Concurrently, studies show that scaling the quantity of fine-tuning data, while maintaining quality, can yield further performance improvements~\cite{scale_1, scale_2, emergent, SFT-quantity-1}.
While prior work has explored dataset quality and size, few studies have examined how a model’s prior knowledge of fine-tuning data influences performance or how different data quantities affect the model’s knowledge. Our study differs by investigating SFT performance on the CBQA task, focusing on how mastery levels and data scale impact model knowledge.

\section{Experiments}
\label{sec:sft}

To explore how SFT affects the factual knowledge of LLMs in the CBQA setting, we conduct a series of controlled experiments. In this section, we outline the datasets used (Section~\ref{sec:dataset}), the models tested (Section~\ref{sec:models}), and the experimental setup (Section~\ref{sec:setup}), followed by a presentation of the results and a summary of our findings (Section~\ref{sec:results}).

\subsection{Dataset}
\label{sec:dataset}

Following~\citet{sft-google} and~\citet{sft-ye}, we use the ENTITYQUESTIONS~\cite{EntityQuestions} to construct the training and testing datasets for our experiments, which is a CBQA-specific dataset containing knowledge across 24 topics extracted from Wikipedia.

\paragraph{Training Data}
Our training dataset, denoted as $\mathcal{D}_{train}$, consists of data on 10 location-related topics extracted from the original training corpus. Following the method of~\citet{sft-ye}, we classify the training samples based on the pre-trained model $\mathcal{M}$’s mastery level on the knowledge associated with each data point $k$. Specifically, we enhance the multi-template completion mechanism of~\citet{sft-ye} to allow $\mathcal{M}$ to complete each data point $k$ using multiple templates. The training data is then divided into five categories according to the proportion $R_k^\mathcal{M}$ of knowledge points correctly completed.\footnote{For additional details on data processing, see Appendix~\ref{sec:detail_data}.} Formally:
\[
\mathcal{D}_{train-i}^\mathcal{M} =
\begin{cases} 
\{k \in \mathcal{D}_{train} \mid R_k^\mathcal{M} = 0\}, \\
~~~~~~~~~~~~~~~~~~~~~~~~~~~~i = 0, \\
\{k \in \mathcal{D}_{train} \mid R_k^\mathcal{M} \in (\frac{i-1}{4}, \frac{i}{4}]\},\\
~~~~~~~~~~~~~~~~~~~~~~~~~~~~i \in \{1, 2, 3, 4\}. 
\end{cases}
\]

\paragraph{Testing Data}
For the in-domain testing dataset \(\mathcal{D}_{test}\), we select data from the same 10 location-related topics in the original test set. Data from the remaining 14 topics are used as the out-of-domain testing dataset \(\mathcal{D}_{testood}\).  
Similar to the training data, both \(\mathcal{D}_{test}\) and \(\mathcal{D}_{testood}\) are categorized as:  
\[
\mathcal{D}_{test} = \bigcup_{i=0}^4 \mathcal{D}_{test-i}^\mathcal{M},~\mathcal{D}_{testood} = \bigcup_{i=0}^4 \mathcal{D}_{testood-i}^\mathcal{M}
\]

An example of data distribution is listed in Table~\ref{tab:data}.\footnote{Data distribution of other LLMs can be found in Appendix~\ref{sec:detail_distribution}.}

\begin{table}[!t]
    \centering
    \resizebox{\linewidth}{!}{
    \begin{tabular}{lccccc}
    \toprule
    \toprule
    \rowcolor{gray!15} {$\mathcal{D}_{train}$} & $\mathcal{D}_{train-0}^\mathcal{M}$ & $\mathcal{D}_{train-1}^\mathcal{M}$ & $\mathcal{D}_{train-2}^\mathcal{M}$ & $\mathcal{D}_{train-3}^\mathcal{M}$ & $\mathcal{D}_{train-4}^\mathcal{M}$ \\ \midrule
    \textbf{Number} & 18456 & 29571 & 11558 & 8923 & 7436 \\ \midrule
    
    \rowcolor{gray!15} $\mathcal{D}_{test}$ & $\mathcal{D}_{test-0}^\mathcal{M}$ & $\mathcal{D}_{test-1}^\mathcal{M}$ & $\mathcal{D}_{test-2}^\mathcal{M}$ & $\mathcal{D}_{test-3}^\mathcal{M}$ & $\mathcal{D}_{test-4}^\mathcal{M}$ \\ \midrule
    \textbf{Number} & 2383 & 3664 & 1484 & 1109 & 915\\ \midrule

    \rowcolor{gray!15} $\mathcal{D}_{testood}$ & $\mathcal{D}_{testood-0}^\mathcal{M}$ & $\mathcal{D}_{testood-1}^\mathcal{M}$ & $\mathcal{D}_{testood-2}^\mathcal{M}$ & $\mathcal{D}_{testood-3}^\mathcal{M}$ & $\mathcal{D}_{testood-4}^\mathcal{M}$ \\ \midrule
         \textbf{Number} & 4127 & 4539 & 1271 & 1120 & 556 \\
         \bottomrule
         \bottomrule
    \end{tabular}
    }
    \caption{An example of data distribution, where $\mathcal{M}$ refers to LLaMA-3-8B.}
    \label{tab:data}
\end{table}

\subsection{Models}
\label{sec:models}

Given the dominance of decoder-only architectures in current LLMs, our analysis focuses exclusively on models of this type. We examine five LLMs from two model families: \textbf{LLaMA-2-7B}, \textbf{LLaMA-2-13B}, and \textbf{LLaMA-2-70B} from the LLaMA-2 family~\cite{LLaMA-2}, and \textbf{LLaMA-3-8B} and \textbf{LLaMA-3-70B} from the LLaMA-3 family~\cite{LLaMA-3}.\footnote{Details of models can be found in Appendix~\ref{sec:detail_model}.}

\begin{figure*}[!t]
    \centering
    \begin{subfigure}[!t]{0.48\linewidth}
        \centering
        \includegraphics[width=\linewidth]{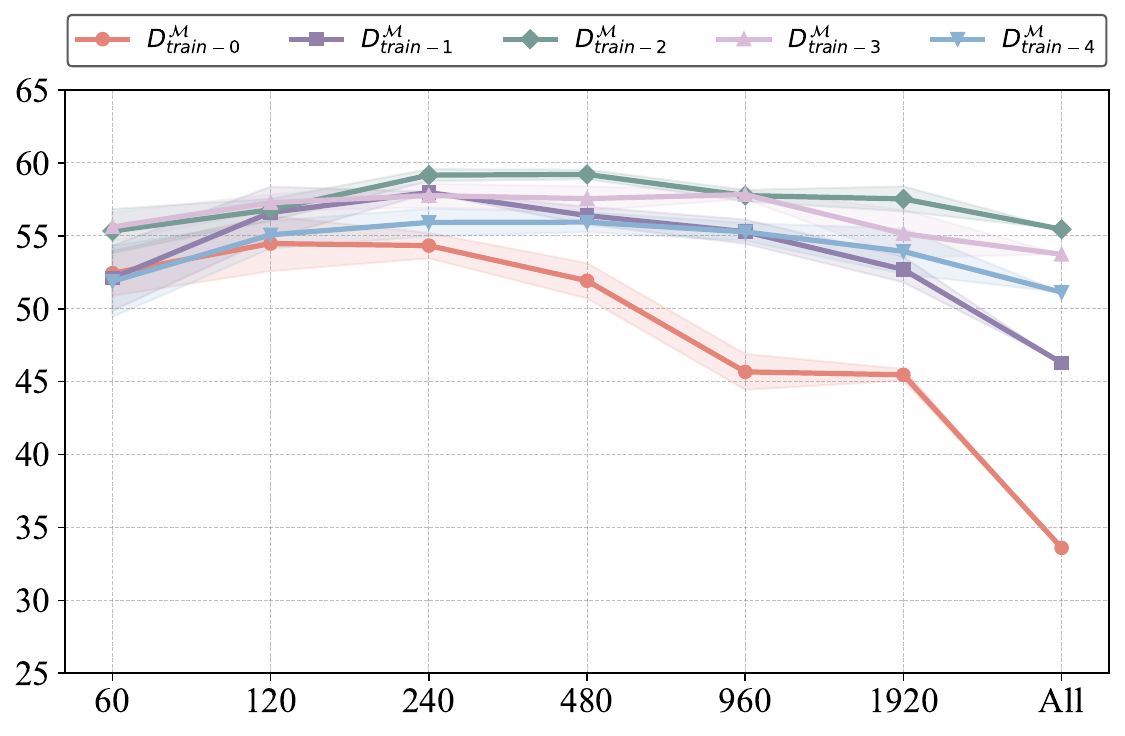}
        \caption{LLaMA-3-8B (In-Domain)}
        \label{fig:result-id-llama-3-8b}
    \end{subfigure}
    \quad
    \begin{subfigure}[!t]{0.48\linewidth}
        \centering
        \includegraphics[width=\linewidth]{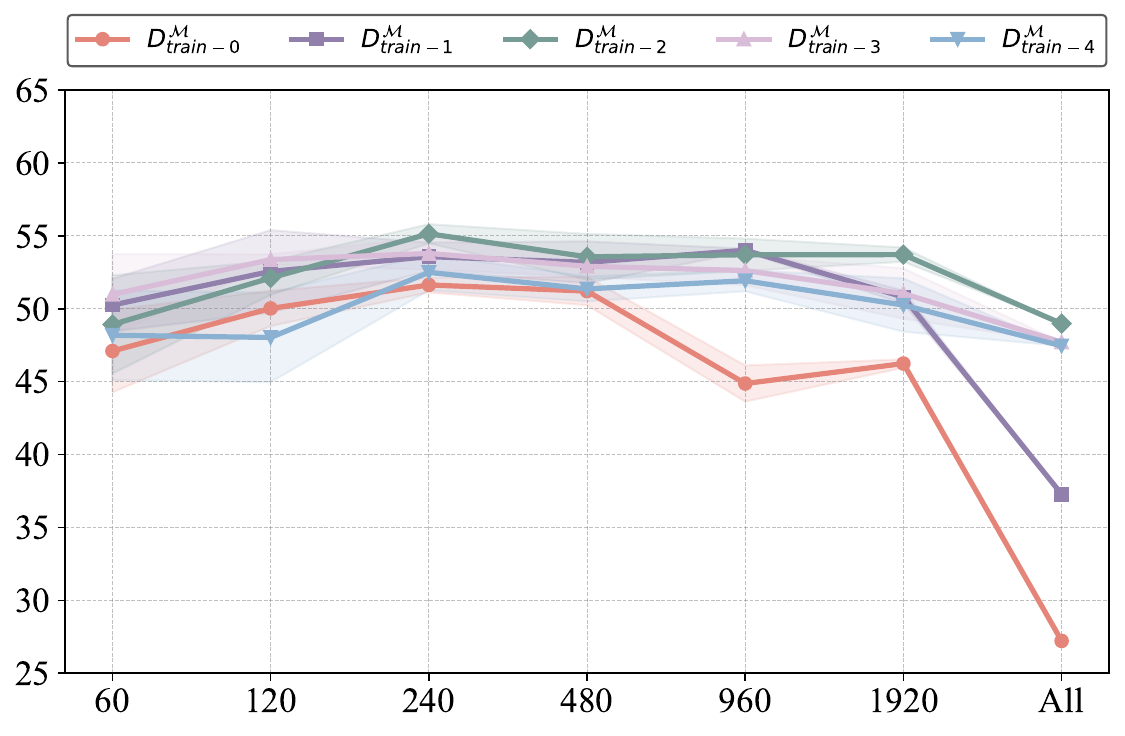}
        \caption{LLaMA-3-8B (Out-of-Domain)}
        \label{fig:result-ood-llama-3-8b}
    \end{subfigure}
    \quad
    \begin{subfigure}[!t]{0.48\linewidth}
        \centering
        \includegraphics[width=\linewidth]{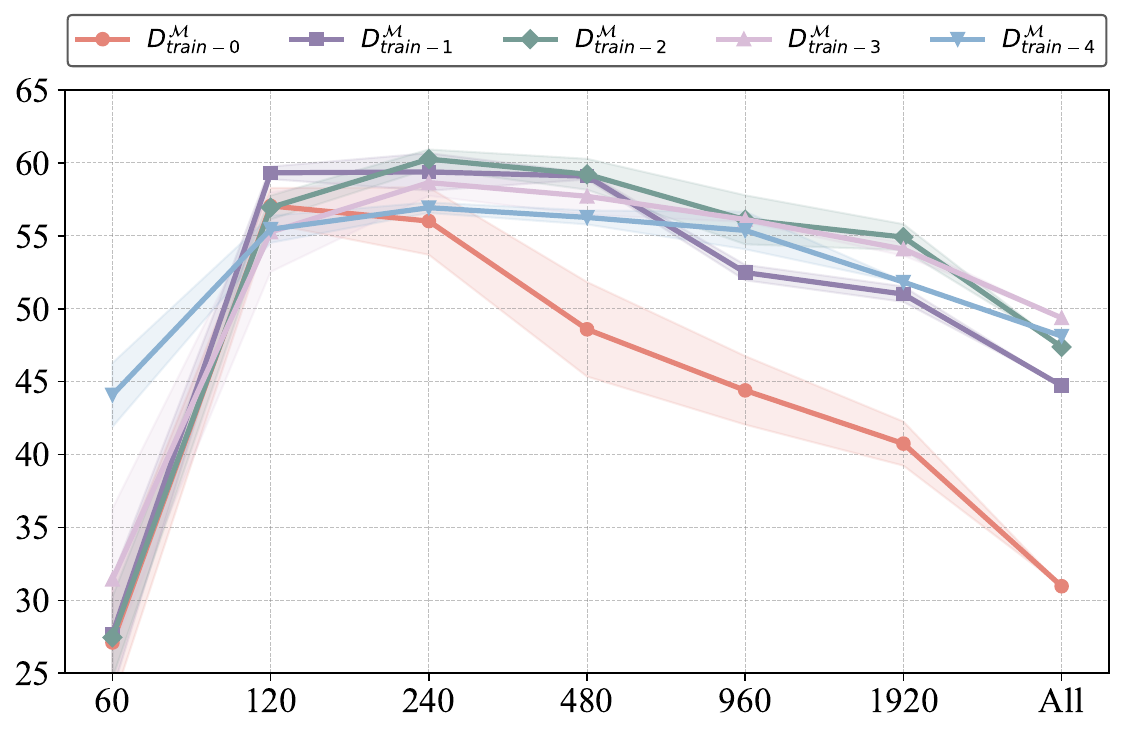}
        \caption{LLaMA-3-70B (In-Domain)}
        \label{fig:result-id-llama-3-70b}
    \end{subfigure}
    \quad
    \begin{subfigure}[!t]{0.48\linewidth}
        \centering
        \includegraphics[width=\linewidth]{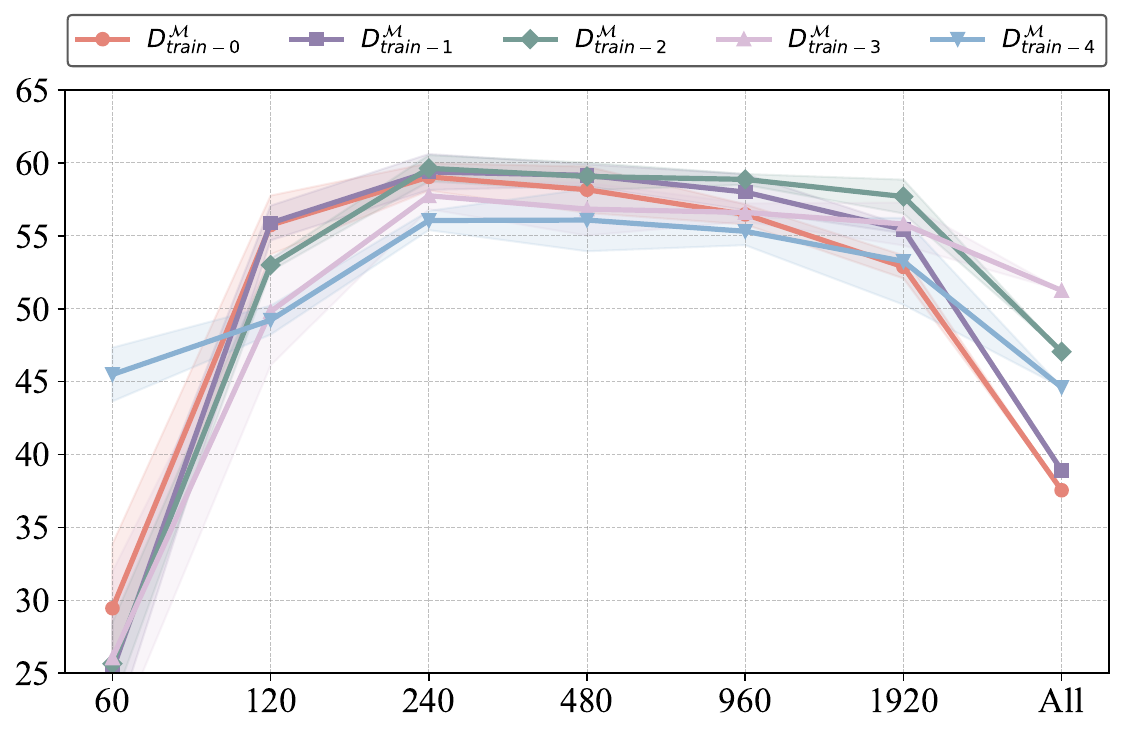}
        \caption{LLaMA-3-70B (Out-of-Domain)}
        \label{fig:result-ood-llama-3-70b}
    \end{subfigure}
    
    \caption{In-domain ($\textbf{Acc}_{test}^\mathcal{M}$) and out-of-domain ($\textbf{Acc}_{testood}^\mathcal{M}$) performance of the LLaMA-3 family models fine-tuned with varying data scales, where `All' indicates the use of the entire dataset listed in Appendix~\ref{sec:detail_distribution}.}
    \label{fig:result-llama-3}
\end{figure*}

\subsection{Experimental Setup}
\label{sec:setup}

Our experiment involves data categorization, training, and testing, aimed at evaluating model performance under diverse settings.

\paragraph{Data Categorization}
To balance the stability and diversity of the generated output, we design 21 mapping templates tailored to each topic's data. The sampling temperature is set to 0.7 to introduce controlled randomness, and each prompt is sampled 10 times to enhance robustness. The output's maximum token length is limited to 32.

\paragraph{Training} Training is conducted using a batch size of 8 over 1 epoch, employing the AdamW~\citep{Adamw} optimizer with cosine learning rate scheduling for stable and efficient convergence. The learning rate is set to $1\times 10^{-5}$.\footnote{To ensure a fair comparison, we use uniform prompt templates during training, as detailed in Appendix~\ref{sec:prompt}.}

\paragraph{Testing}
For testing, we utilize a greedy decoding strategy with a maximum output length of 16, maintaining consistency with the prompt templates used during training. To mitigate bias from the training data selection, we generate five distinct training datasets by random sampling. Each experiment is repeated using these datasets, and the final results are reported as the mean and variance across the five runs. Evaluation metrics include accuracy, categorized by different knowledge mastery levels, with the mean accuracy across all test sets serving as the final metric:
\[
\textbf{Acc}_{test}^{\mathcal{M}} = \sum_{i=0}^4 \textbf{Acc}_{test-i}^{\mathcal{M}}/5
\]
\[
\textbf{Acc}_{testood}^{\mathcal{M}} = \sum_{i=0}^4 \textbf{Acc}_{testood-i}^{\mathcal{M}}/5
\]

\subsection{Main Results}
\label{sec:results}

\begin{table*}[!t]
    \centering
    \resizebox{\linewidth}{25mm}{
    \begin{tabular}{lcccccccccccc}
    \toprule
    \multirow{2}*{\textbf{Source}} & \multicolumn{6}{c}{\textbf{In-Domain}} & \multicolumn{6}{c}{\textbf{Out-of-Domain}} \\ \cmidrule(lr){2-7} \cmidrule(lr){8-13}
    & $\textbf{Acc}_{test-0}^\mathcal{M}$ & $\textbf{Acc}_{test-1}^\mathcal{M}$ & $\textbf{Acc}_{test-2}^\mathcal{M}$ & $\textbf{Acc}_{test-3}^\mathcal{M}$ & $\textbf{Acc}_{test-4}^\mathcal{M}$ & $\textbf{Acc}_{test}^\mathcal{M}$ & $\textbf{Acc}_{testood-0}^\mathcal{M}$ & $\textbf{Acc}_{testood-1}^\mathcal{M}$ & $\textbf{Acc}_{testood-2}^\mathcal{M}$ & $\textbf{Acc}_{testood-3}^\mathcal{M}$ & $\textbf{Acc}_{testood-4}^\mathcal{M}$ & $\textbf{Acc}_{testood}^\mathcal{M}$ \\ \midrule
    \rowcolor{gray!15} \multicolumn{13}{c}{$\mathcal{M}=\textit{LLaMA-3-8B}$} \\
    $\mathcal{D}_{train-0}^\mathcal{M}$ & $\mathbf{1.75}_{0.17}$ & $16.07_{0.67}$ & $55.03_{1.39}$ & $71.06_{1.09}$ & $83.46_{1.23}$ & $45.47_{0.40}$ & $\mathbf{1.91}_{0.33}$ & $15.89_{1.20}$ & $59.01_{0.51}$ & $74.08_{0.63}$ & $80.33_{0.98}$ & $46.24_{0.29}$ \\
    $\mathcal{D}_{train-1}^\mathcal{M}$ & $0.98_{0.14}$ & $\mathbf{40.12}_{0.74}$ & $63.93_{0.55}$ & $74.19_{0.73}$ & $84.22_{3.96}$ & $52.69_{0.88}$ & $1.66_{0.09}$ & $23.88_{0.45}$ & $65.03_{0.77}$ & $79.63_{0.63}$ & $83.84_{0.55}$ & $50.80_{0.45}$ \\
    $\mathcal{D}_{train-2}^\mathcal{M}$ & $0.78_{0.03}$ & $36.56_{0.53}$ & $\mathbf{75.61}_{1.18}$ & $83.98_{1.37}$ & $90.71_{1.31}$ & $\mathbf{57.53}_{0.86}$ & $1.45_{0.35}$ & $\mathbf{25.02}_{0.30}$ & $\mathbf{70.52}_{1.59}$ & $\mathbf{83.66}_{0.67}$ & $87.89_{0.45}$ & $\mathbf{53.71}_{0.49}$ \\
    $\mathcal{D}_{train-3}^\mathcal{M}$ & $0.64_{0.15}$ & $27.20_{3.69}$ & $70.33_{1.73}$ & $\mathbf{85.90}_{1.47}$ & $91.66_{1.57}$ & $55.15_{1.64}$ & $1.39_{0.34}$ & $21.66_{3.13}$ & $63.91_{2.70}$ & $81.34_{0.93}$ & $86.87_{1.85}$ & $51.04_{1.73}$ \\
    $\mathcal{D}_{train-4}^\mathcal{M}$ & $0.64_{0.06}$ & $24.26_{3.38}$ & $68.28_{2.00}$ & $83.29_{1.23}$ & $\mathbf{93.19}_{1.91}$ & $53.93_{1.56}$ & $0.93_{0.11}$ & $17.72_{1.33}$ & $63.64_{4.39}$ & $80.55_{2.05}$ & $\mathbf{88.43}_{1.47}$ & $50.25_{1.83}$ \\
    \midrule
    \rowcolor{gray!15} \multicolumn{13}{c}{$\mathcal{M}=\textit{LLaMA-3-70B}$} \\
    $\mathcal{D}_{train-0}^\mathcal{M}$ & $\mathbf{3.72}_{0.33}$ & $22.68_{1.53}$ & $47.28_{1.26}$ & $57.97_{2.25}$ & $72.08_{3.20}$ & $40.75_{1.51}$ & $\mathbf{3.08}_{0.39}$ & $25.90_{1.59}$ & $67.04_{1.63}$ & $82.61_{0.95}$ & $85.74_{1.30}$ & $52.87_{0.79}$ \\
    $\mathcal{D}_{train-1}^\mathcal{M}$ & $1.94_{0.11}$ & $\mathbf{43.85}_{0.29}$ & $63.45_{1.47}$ & $66.22_{1.66}$ & $79.54_{0.65}$ & $51.00_{0.53}$ & $2.61_{0.45}$ & $31.01_{0.79}$ & $72.63_{0.16}$ & $84.69_{0.30}$ & $86.22_{0.69}$ & $55.43_{0.26}$ \\
    $\mathcal{D}_{train-2}^\mathcal{M}$ & $1.23_{0.07}$ & $38.17_{1.78}$ & $\mathbf{71.68}_{0.82}$ & $77.58_{1.27}$ & $85.89_{1.44}$ & $\mathbf{54.91}_{0.89}$ & $2.06_{0.50}$ & $\mathbf{31.26}_{2.10}$ & $\mathbf{74.51}_{1.27}$ & $88.63_{0.97}$ & $92.01_{1.19}$ & $\mathbf{57.69}_{1.16}$ \\
    $\mathcal{D}_{train-3}^\mathcal{M}$ & $1.00_{0.11}$ & $31.52_{0.61}$ & $68.32_{0.30}$ & $\mathbf{81.11}_{0.73}$ & $88.49_{1.60}$ & $54.09_{0.45}$ & $1.91_{0.79}$ & $26.70_{1.71}$ & $69.60_{2.77}$ & $\mathbf{89.61}_{1.44}$ & $91.22_{1.39}$ & $55.81_{1.47}$ \\
    $\mathcal{D}_{train-4}^\mathcal{M}$ & $0.90_{0.05}$ & $26.16_{1.45}$ & $64.27_{0.75}$ & $78.00_{0.43}$ & $\mathbf{89.83}_{0.77}$ & $51.83_{0.05}$ & $0.81_{0.35}$ & $21.80_{3.65}$ & $66.52_{5.65}$ & $84.85_{2.57}$ & $\mathbf{92.29}_{2.63}$ & $53.25_{2.97}$ \\
    \bottomrule
    \end{tabular}
    }
    \caption{Performance of the fine-tuned LLaMA-3 family models on in-domain and out-of-domain test sets, using 1920 data points with varying levels of mastery.}
    \label{tab:1920-detail}
\end{table*}

We fine-tune each of the five selected LLMs using datasets with five different mastery levels. To conduct a more detailed analysis, we compare changes in model performance across varying data scales. To enhance robustness, we ensure a balanced data distribution across topics and repeat each experiment three times. Figure~\ref{fig:result-llama-3} presents the in-domain and out-of-domain test results for the LLaMA-3 family of models.\footnote{Test results for the LLaMA-2 family of models can be found in Appendix~\ref{sec:result-llama2}.}
From the results, we observe two unexpected phenomena.

\paragraph{\textit{Phenomenon 1}}
\textit{Regardless of the type of training data used, LLMs achieve their optimal performance with just 240 data points. Adding more training data beyond this point risks degrading model performance.}

Our analysis reveals that model performance improves as the amount of fine-tuned data increases from 60 to 240 entries, aligning with the general expectation that more data enhances performance. However, performance peaks at \textbf{only 240 entries}, and adding additional fine-tuned data not only fails to yield further improvements but often leads to a significant decline. For instance, when fine-tuned with barely mastered data (i.e., \(\mathcal{D}_{train-0}^\mathcal{M}\)), LLaMA-3-8B achieves an \(\textbf{Acc}_{test}^\mathcal{M}\) score that is 8.86\% lower when trained with 1,920 entries compared to 240 entries. A decline of 13.69\% is even observed when comparing 240 entries from \(\mathcal{D}_{train-2}^\mathcal{M}\). Notably, when LLMs are trained with the full dataset for each data category, their performance on the CBQA task is nearly at its lowest across all data categories. This striking finding suggests that increasing the volume of fine-tuned data does not necessarily enhance model knowledge and may impair it.

\begin{figure}[!t]
    \centering
        \includegraphics[width=\linewidth]{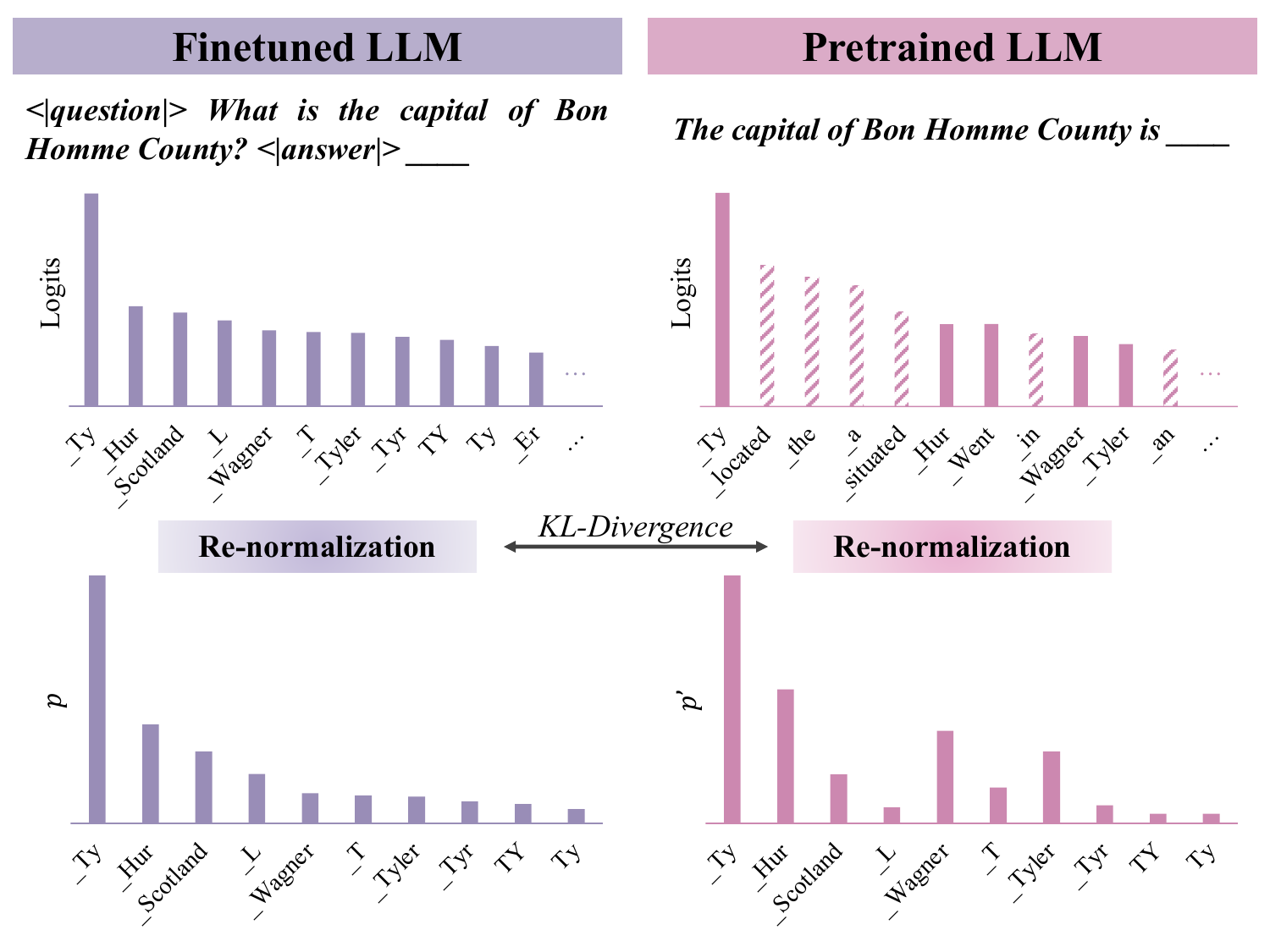}
    \caption{Illustration of logits re-normalization. Since the pre-trained LLM tends to assign high probabilities to common dummy words, we identify the ten highest logits in the fine-tuned LLM and extract the corresponding values from the pre-trained LLM. After re-normalization, we compute the KL divergence to quantify the distributional difference.}
    \label{fig:logits}
\end{figure}

\paragraph{\textit{Phenomenon 2}}
\textit{When the amount of fine-tuned data reaches a certain threshold (e.g., 1,920 entries), model performance varies significantly based on the knowledge mastery level of the training data.}

While model performance generally declines when the fine-tuned data exceeds 240 entries, the rate of decline differs depending on the knowledge mastery level of the training data. Notably, models fine-tuned with data from \(\mathcal{D}_{train-0}^\mathcal{M}\) exhibit a steeper performance drop compared to those trained on other data types. For instance, when fine-tuned with 1,920 entries, the \(\textbf{Acc}_{test}^\mathcal{M}\) difference between LLaMA-3-8B models trained on \(\mathcal{D}_{train-0}^\mathcal{M}\) and \(\mathcal{D}_{train-2}^\mathcal{M}\) reaches 12.06\%, which is 1.50 times the difference observed with only 240 training entries. Table~\ref{tab:1920-detail} illustrates the performance of LLaMA-3 family models across various test sets when fine-tuned with 1,920 entries from different categories. The results show that models trained on \(\mathcal{D}_{train-0}^\mathcal{M}\) experience substantial performance degradation on test sets other than \(\mathcal{D}_{test-0}^\mathcal{M}\). More generally, training on low-mastery data significantly impairs performance on high-mastery test data. Conversely, training on high-mastery data (e.g., \(\mathcal{D}_{train-4}^\mathcal{M}\)) leads to suboptimal performance on low-mastery test data. Training with mid-level mastery data, such as \(\mathcal{D}_{train-2}^\mathcal{M}\), strikes a better balance, yielding superior overall performance.

\section{Token-Level Analysis}
\label{sec:token-level}

To explain the performance variation observed across fine-tuned LLMs, we analyze how fine-tuning alters token-level output distributions compared to the pre-trained model.
Specifically, we compute the divergence in predicted token distributions between fine-tuned and pre-trained models using KL divergence (Section~\ref{sec:kl}). This token-level analysis reveals some interesting findings (Section~\ref{sec:token-results}).

\subsection{KL Divergence Computation}
\label{sec:kl}

\begin{figure}[!t]
    \centering
        \includegraphics[width=\linewidth]{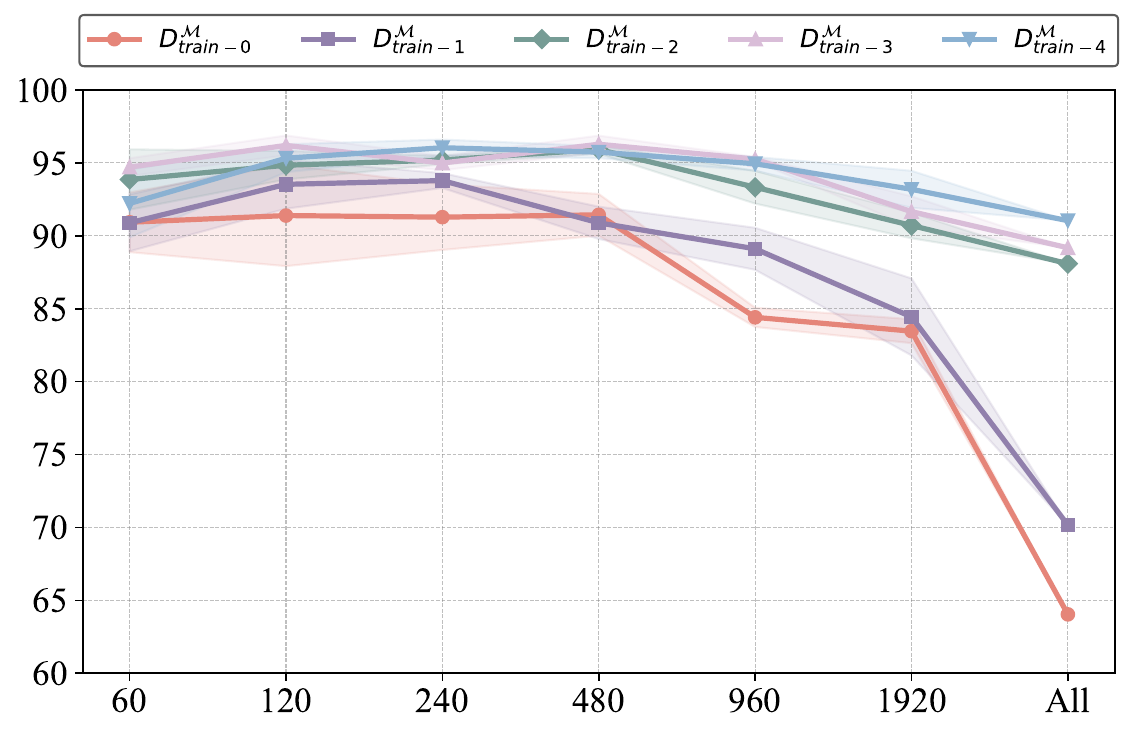}
    \caption{Performance on \(\mathcal{D}_{test-4}^\mathcal{M}\) (\(\textbf{Acc}_{test-4}^\mathcal{M}\)) of LLMs fine-tuned on LLaMA-3-8B.}
    \label{fig:result-test-4}
\end{figure}

Given the performance degradation observed in Section~\ref{sec:results}, we investigate the underlying token distribution shifts caused by SFT. Specifically, we use KL divergence to quantify the differences in token probabilities between fine-tuned and pre-trained models. A higher KL divergence suggests a more significant shift in the model’s token probability distribution.

\paragraph{Data Selection}
Given that the pre-trained model is used to complement the prior text, the quality of its completions depends on both the input prompt and the structure of the mapping template, as outlined in Section~\ref{sec:setup}. The selection of appropriate data is critical to ensuring the robustness of the results. For \(\mathcal{D}_{test-4}^\mathcal{M}\), we observe that the pre-trained model's completion success rate exceeds 75\% across multiple samples and templates, suggesting that this dataset is relatively insensitive to variations in the mapping template. In contrast, other datasets are more sensitive to such variations, so our comparison of different LLMs in this section is limited to \(\mathcal{D}_{test-4}^\mathcal{M}\). For each topic, we select the mapping template yielding the highest success rate across samples and focus our analysis on tokens in completions where the answers appear near the beginning of the generated text.

\paragraph{Logits Re-normalization}
Our goal is to compute the KL divergence between the logits distributions for the first token predicted by both the fine-tuned and pre-trained LLMs. However, as shown in Figure~\ref{fig:logits}, the pre-trained model tends to assign higher probabilities to common dummy words (e.g., `the', `a', etc.), whereas fine-tuned models typically reduce the likelihood of these words in favor of more relevant tokens. If we directly compute the KL divergence on the raw logits, these dummy words could distort the results and obscure meaningful differences between the models.
To mitigate this issue, we introduce a logits re-normalization procedure. Specifically, we sort the logits predicted by the fine-tuned model and extract the top 10 values, denoted as \( l_0, l_1, \dots, l_9 \). We then identify the corresponding logits, \( l_0', l_1', \dots, l_9' \), from the pre-trained model’s completions. Moreover, we apply the softmax function to these logits to derive their normalized probabilities, respectively:
\[
p_i = \text{Softmax}(l_i),~p_i' = \text{Softmax}(l_i').
\]

After completing the logits re-normalization, we compute the KL divergence between the probability distributions \( p \) and \( p' \) for the fine-tuned and pre-trained models as follows:
\[
s_{\text{KL}}(p \parallel p') = -\sum_i p_i \log \frac{p_i'}{p_i}.
\]

\subsection{Results Analysis}
\label{sec:token-results}

\begin{figure}[!t]
    \centering
        \includegraphics[width=\linewidth]{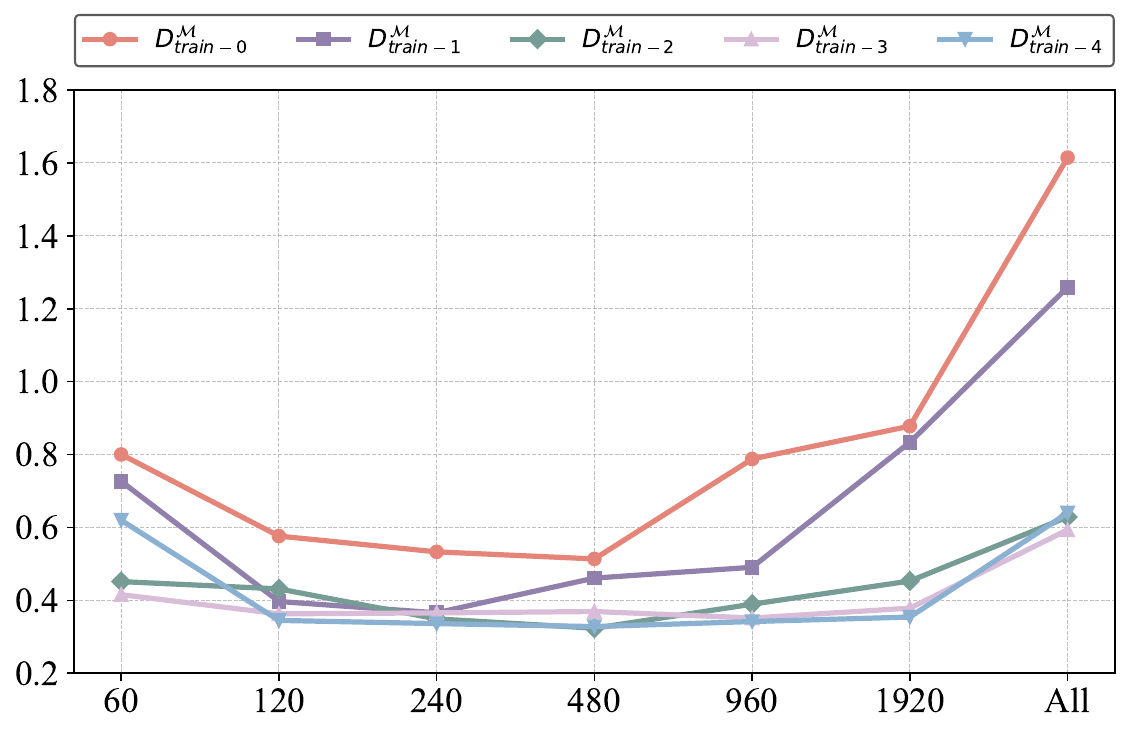}
    \caption{KL divergence of logits distribution between LLaMA-3-8B fine-tuned with different datasets and the pre-trained one.}
    \label{fig:logits_kl}
\end{figure}

We analyze the performance of individual LLMs fine-tuned based on LLaMA-3-8B, presenting their results on \( \mathcal{D}_{test-4}^\mathcal{M} \) in Figure~\ref{fig:result-test-4} and their KL divergence relative to the pre-trained model's distribution in Figure~\ref{fig:logits_kl}. From these results, we derive two key findings.  

\paragraph{\textit{Finding 1}}
\textit{Regardless of the category of fine-tuning data, the difference in predicted logits distributions between the fine-tuned and pre-trained models initially decreases and then increases as the amount of data grows.}  

Figure~\ref{fig:logits_kl} illustrates how the predicted logits distributions of fine-tuned model diverge from the pre-trained model as training data increases. When fine-tuning with a small dataset (e.g., 60 samples), the logits distribution shifts significantly due to insufficient data, leading to unstable training. As the dataset grows (e.g., 240 samples), this discrepancy decreases, indicating improved stability. However, with further increases, the difference in logits distributions grows again, particularly for models trained on \( \mathcal{D}_{train-0}^\mathcal{M} \) and \( \mathcal{D}_{train-1}^\mathcal{M} \). This suggests that as training data increases, the model deviates further from its pre-trained knowledge. The effect is more pronounced when fine-tuning on low-mastery data, making the model more susceptible to knowledge shifts.  

\paragraph{\textit{Finding 2}}
\textit{As the difference in the predicted logits distribution between the fine-tuned model and the pre-trained model increases, model performance declines, indicating a negative impact of excessive knowledge shifts.}  

Figure~\ref{fig:result-test-4} and Figure~\ref{fig:logits_kl} reveal a strong correlation between performance degradation on \( \mathcal{D}_{test-4}^\mathcal{M} \) and increasing divergence in logits distributions. Since \( \mathcal{D}_{test-4}^\mathcal{M} \) contains samples well mastered by the pre-trained model, substantial shifts in learned knowledge during fine-tuning can lead to catastrophic forgetting, where previously acquired knowledge is lost, thereby degrading performance.  
This effect is particularly evident when training with large datasets. For instance, the model fine-tuned on \( \mathcal{D}_{train-0}^\mathcal{M} \) experiences the most significant knowledge shift and performs the worst among all fine-tuned models. Since changes in logits distribution reflect underlying modifications to model parameters, we hypothesize that \textbf{{excessive parameter updates} during fine-tuning, especially when using large or low-mastery datasets, lead to overall performance decline}.

\section{Parameter-Level Analysis}
\label{sec:parameter-level}

The observations and analyses in Section~\ref{sec:token-level} indicate that excessive parameter updates can degrade model performance. To further investigate this, we analyze the impact at the parameter level by progressively restoring the updated parameters and examining the resulting performance changes (Section~\ref{sec:restoring}). Our findings indicate that a significant proportion of parameter updates during SFT do not contribute to performance improvement and may even be detrimental (Section~\ref{sec:param-results}).

\subsection{Parameter Restoration}
\label{sec:restoring}

To examine the impact of excessive parameter updates on model performance, we design an experimental framework for parameter restoration. Specifically, we compare the fine-tuned model with the pre-trained model, sorted by the rate of parameter change.\footnote{Specific calculation details can be found in Appendix~\ref{sec:restore-calcu}.} Table~\ref{tab:rank} reports the percentage of total parameter updates attributed to different proportions of the most highly updated parameters in LLMs fine-tuned on LLaMA-3-8B. The results indicate that parameter updates are heavily concentrated in a small subset of parameters. For instance, more than 70\% of the total updates occur in fewer than 1\% of the parameters.
Following this, we progressively restore the most significantly updated parameters to their original values in the pre-trained model, starting with the largest updates and gradually including smaller ones, while monitoring the corresponding changes in model performance. This process is illustrated in Figure~\ref{fig:illustration-restore}.

\begin{table}[!t]
    \centering
    \resizebox{\linewidth}{!}
    {
    \begin{tabular}{lcccccccc}
    \toprule
    \textbf{Proportion} & \textbf{1\%} & \textbf{3\%} & \textbf{5\%} & \textbf{10\%} & \textbf{20\%} & \textbf{40\%} & \textbf{60\%} \\ \midrule
    \rowcolor{gray!15} \multicolumn{8}{c}{\textit{Number of Training Data: 240}} \\
    $\mathcal{D}_{train-0}^\mathcal{M}$ & 70.59\% & 78.82\% & 82.35\% & 87.06\% & 91.76\% & 96.47\% & 99.12\% \\
    $\mathcal{D}_{train-1}^\mathcal{M}$ & 71.01\% & 79.29\% & 82.84\% & 87.57\% & 92.31\% & 97.04\% & 99.11\% \\
    $\mathcal{D}_{train-2}^\mathcal{M}$ & 71.13\% & 79.17\% & 82.74\% & 87.50\% & 92.26\% & 96.43\% & 99.12\% \\
    $\mathcal{D}_{train-3}^\mathcal{M}$ & 70.72\% & 78.97\% & 82.51\% & 87.22\% & 91.93\% & 96.65\% & 99.09\% \\
    $\mathcal{D}_{train-4}^\mathcal{M}$ & 70.98\% & 78.74\% & 82.18\% & 87.36\% & 91.95\% & 96.55\% & 99.04\% \\\midrule
    \rowcolor{gray!15} \multicolumn{8}{c}{\textit{Number of Training Data: 1920}} \\
    $\mathcal{D}_{train-0}^\mathcal{M}$ & 70.56\% & 78.50\% & 82.24\% & 86.92\% & 92.06\% & 96.26\% & 98.69\% \\
    $\mathcal{D}_{train-1}^\mathcal{M}$ & 70.89\% & 78.87\% & 82.63\% & 87.32\% & 92.02\% & 96.71\% & 98.69\% \\
    $\mathcal{D}_{train-2}^\mathcal{M}$ & 70.75\% & 78.77\% & 82.08\% & 87.26\% & 91.98\% & 96.70\% & 98.70\% \\
    $\mathcal{D}_{train-3}^\mathcal{M}$ & 70.74\% & 78.70\% & 81.98\% & 87.13\% & 91.82\% & 96.50\% & 98.70\% \\
    $\mathcal{D}_{train-4}^\mathcal{M}$ & 70.83\% & 78.70\% & 82.41\% & 87.04\% & 92.13\% & 96.30\% & 98.70\% \\\bottomrule
    \end{tabular}
    }
    \caption{Percentage of total parameter updates concentrated in different proportions of the most highly updated parameters in various LLMs fine-tuned on LLaMA-3-8B.}
    \label{tab:rank}
\end{table}

\subsection{Results Analysis}
\label{sec:param-results}

\begin{table*}[!t]
    \centering
    \begin{subtable}[!t]{0.48\linewidth}
    \centering
    \resizebox{\linewidth}{!}
    {
    \begin{tabular}{lccccc}
    \toprule
    \textbf{Restore}  & $\mathbf{\mathcal{D}_{train-0}^\mathcal{M}}$ & $\mathbf{\mathcal{D}_{train-1}^\mathcal{M}}$ & $\mathbf{\mathcal{D}_{train-2}^\mathcal{M}}$ & $\mathbf{\mathcal{D}_{train-3}^\mathcal{M}}$ & $\mathbf{\mathcal{D}_{train-4}^\mathcal{M}}$ \\ \midrule
    \rowcolor{gray!15} \multicolumn{6}{c}{\textit{Number of Training Data: 240}} \\
    0 & 55.33 & 57.96 & 59.32 & 59.12 & 53.97 \\
        1\% & \cellcolor{GREEN!10}55.76 & \cellcolor{GREEN!10}58.17 & \cellcolor{GREEN!10}59.62 & \cellcolor{GREEN!10}59.24 & \cellcolor{GREEN!10}54.30 \\
    3\% & \cellcolor{GREEN!20}56.64 & \cellcolor{GREEN!10}58.52 & \cellcolor{GREEN!10}59.77 & \cellcolor{GREEN!10}59.40 & \cellcolor{GREEN!10}54.31 \\
    5\% & \cellcolor{GREEN!20}57.22 & \cellcolor{GREEN!10}58.68 & \cellcolor{GREEN!10}59.89 & \cellcolor{GREEN!10}59.63 & \cellcolor{GREEN!10}54.44 \\
    10\% & \cellcolor{GREEN!30}58.32 & \cellcolor{GREEN!20}59.45 & \cellcolor{GREEN!20}60.40 & \cellcolor{GREEN!10}59.83 & \cellcolor{GREEN!10}54.69 \\
    20\% & \cellcolor{GREEN!40}59.07 & \cellcolor{GREEN!20}59.81 & \cellcolor{GREEN!10}59.88 & \cellcolor{GREEN!10}59.91 & \cellcolor{RED!7}46.45 \\
    40\% & \cellcolor{GREEN!40}59.77 & \cellcolor{RED!24}33.40 & \cellcolor{RED!17}42.44 & \cellcolor{RED!48}11.20 & \cellcolor{RED!30}23.83 \\
    60\% & \cellcolor{RED!54}1.68 & \cellcolor{RED!55}2.20 & \cellcolor{RED!56}3.65 & \cellcolor{RED!57}2.56 & \cellcolor{RED!52}1.65 \\\midrule
    \rowcolor{gray!15} \multicolumn{6}{c}{\textit{Number of Training Data: 1920}} \\
    0 & 44.96 & 52.43 & 58.80 & 57.70 & 55.22 \\
    1\% & \cellcolor{GREEN!20}46.73 & \cellcolor{GREEN!10}53.72 & \cellcolor{GREEN!15}59.85 & \cellcolor{GREEN!15}58.68 & \cellcolor{GREEN!8}55.88 \\
    3\% & \cellcolor{GREEN!40}48.53 & \cellcolor{GREEN!30}55.01 & \cellcolor{GREEN!20}60.56 & \cellcolor{GREEN!20}59.23 & \cellcolor{GREEN!10}56.76 \\
    5\% & \cellcolor{GREEN!50}49.85 & \cellcolor{GREEN!35}55.96 & \cellcolor{GREEN!25}61.10 & \cellcolor{GREEN!25}59.65 & \cellcolor{GREEN!20}57.34 \\
    10\% & \cellcolor{GREEN!75}52.10 & \cellcolor{GREEN!50}57.14 & \cellcolor{GREEN!30}61.67 & \cellcolor{GREEN!30}60.02 & \cellcolor{GREEN!30}58.24 \\
    20\% & \cellcolor{GREEN!95}54.81 & \cellcolor{GREEN!60}58.33 & \cellcolor{GREEN!40}62.21 & \cellcolor{GREEN!10}58.93 & \cellcolor{GREEN!30}58.66 \\
    40\% & \cellcolor{GREEN}55.44 & \cellcolor{RED!30}22.06 & \cellcolor{GREEN!10}59.97 & \cellcolor{RED!51}6.92 & \cellcolor{GREEN!10}56.50 \\
    60\% & \cellcolor{RED!43}1.48 & \cellcolor{RED!51}1.12 & \cellcolor{RED!57}1.62 & \cellcolor{RED!57}0.51 & \cellcolor{RED!54}0.60 \\\bottomrule
    \end{tabular}
    }
    \caption{In-Domain ($\textbf{Acc}_{test}^\mathcal{M}$)}
    \end{subtable}
    \quad
    \begin{subtable}[!t]{0.48\linewidth}
    \centering
    \resizebox{\linewidth}{!}
    {
    \begin{tabular}{lccccc}
    \toprule
    \textbf{Restore}  & $\mathbf{\mathcal{D}_{train-0}^\mathcal{M}}$ & $\mathbf{\mathcal{D}_{train-1}^\mathcal{M}}$ & $\mathbf{\mathcal{D}_{train-2}^\mathcal{M}}$ & $\mathbf{\mathcal{D}_{train-3}^\mathcal{M}}$ & $\mathbf{\mathcal{D}_{train-4}^\mathcal{M}}$ \\ \midrule
    \rowcolor{gray!15} \multicolumn{6}{c}{\textit{Number of Training Data: 240}} \\
    0 & 52.37 & 51.70 & 55.35 & 55.23 & 50.69 \\
    1\% & \cellcolor{GREEN!5}52.62 & \cellcolor{GREEN!10}52.39 & \cellcolor{GREEN!10}56.45 & \cellcolor{GREEN!10}56.17 & \cellcolor{GREEN!5}50.82 \\
    3\% & \cellcolor{GREEN!10}53.03 & \cellcolor{GREEN!10}52.82 & \cellcolor{GREEN!10}56.47 & \cellcolor{GREEN!10}56.41 & \cellcolor{GREEN!5}50.74 \\
    5\% & \cellcolor{GREEN!10}53.27 & \cellcolor{GREEN!20}53.09 & \cellcolor{GREEN!15}56.80 & \cellcolor{GREEN!13}56.56 & \cellcolor{RED!1}50.59 \\
    10\% & \cellcolor{GREEN!11}53.44 & \cellcolor{GREEN!21}53.87 & \cellcolor{GREEN!11}56.46 & \cellcolor{GREEN!15}56.72 & \cellcolor{RED!1}49.71 \\
    20\% & \cellcolor{GREEN!18}54.18 & \cellcolor{GREEN!26}54.36 & \cellcolor{GREEN!10}55.95 & \cellcolor{GREEN!5}55.52 & \cellcolor{RED!7}43.13 \\
    40\% & \cellcolor{GREEN!14}53.79 & \cellcolor{RED!31}20.77 & \cellcolor{RED!10}45.49 & \cellcolor{RED!38}17.56 & \cellcolor{RED!19}31.19 \\
    60\% & \cellcolor{RED!52}0.20 & \cellcolor{RED!51}0.22 & \cellcolor{RED!55}0.32 & \cellcolor{RED!55}0.20 & \cellcolor{RED!50}0.23 \\\midrule
    \rowcolor{gray!15} \multicolumn{6}{c}{\textit{Number of Training Data: 1920}} \\
    0 & 49.40 & 52.38 & 54.04 & 53.79 & 51.70 \\
    1\% & \cellcolor{GREEN!13}50.78 & \cellcolor{GREEN!20}54.20 & \cellcolor{GREEN!11}55.17 & \cellcolor{GREEN!10}54.75 & \cellcolor{GREEN!9}52.62 \\
    3\% & \cellcolor{GREEN!26}52.03 & \cellcolor{GREEN!30}55.12 & \cellcolor{GREEN!20}56.00 & \cellcolor{GREEN!20}55.52 & \cellcolor{GREEN!16}53.35 \\
    5\% & \cellcolor{GREEN!31}52.54 & \cellcolor{GREEN!28}55.12 & \cellcolor{GREEN!25}56.34 & \cellcolor{GREEN!21}55.84 & \cellcolor{GREEN!20}53.77 \\
    10\% & \cellcolor{GREEN!40}53.42 & \cellcolor{GREEN!27}55.08 & \cellcolor{GREEN!26}56.68 & \cellcolor{GREEN!20}55.54 & \cellcolor{GREEN!26}54.32 \\
    20\% & \cellcolor{GREEN!51}54.50 & \cellcolor{GREEN!16}53.91 & \cellcolor{GREEN!31}57.10 & \cellcolor{RED!1}52.23 & \cellcolor{GREEN!21}53.82 \\
    40\% & \cellcolor{GREEN!42}53.64 & \cellcolor{RED!32}20.51 & \cellcolor{RED!1}53.84 & \cellcolor{RED!44}9.67 & \cellcolor{RED!1}50.17 \\
    60\% & \cellcolor{RED!49}0.30 & \cellcolor{RED!53}0.10 & \cellcolor{RED!54}0.27 & \cellcolor{RED!53}0.07 & \cellcolor{RED!51}0.18 \\\bottomrule
    \end{tabular}
    }
    \caption{Out-of-Domain ($\textbf{Acc}_{testood}^\mathcal{M}$)}
    \end{subtable}
    \caption{Performance of LLaMA-3-8B after restoring different scales of parameters across various fine-tuning datasets. Improvements over the non-restored model are highlighted in \colorbox{GREEN}{green}, while performance declines are shown in \colorbox{RED}{red}, with darker shades indicating larger differences.}
    \label{tab:restore}
\end{table*}

We evaluate the performance of LLaMA-3-8B after restoring different proportions of parameters across various fine-tuning datasets. The results are summarized in Table~\ref{tab:restore}.
Our analysis of these results reveals several noteworthy findings.

\begin{table}[!t]
    \centering
    \resizebox{0.9\linewidth}{!}
    {
    \begin{tabular}{lcccc}
    \toprule
    \multirow{2}*{\textbf{Restore}} & \multicolumn{3}{c}{\textbf{XSum}} & \textbf{GSM8K} \\ \cmidrule(lr){2-4} \cmidrule(lr){5-5}
    & \textbf{ROUGE-1} & \textbf{ROUGE-2} & \textbf{ROUGE-L} & \textbf{ACC} \\ \midrule
    0 & 42.57 & 19.50 & 34.55 & 57.69  \\
    1\% & 42.50 & 19.71 & 34.67 & 57.69\\
    3\% & \textbf{42.63} & \textbf{19.78} & \textbf{34.75} & 57.75 \\
    5\% & 42.36 & 19.47 & 34.44 & 58.49 \\
    10\% & 42.57 & 19.40 & 34.60 & \textbf{59.60} \\
    20\% & 41.31 & 18.59 & 33.51 & 58.72 \\
    40\% & 15.59 & 4.15 & 12.09 & 0 \\
    60\% & 0 & 0 & 0 & 0 \\
    \bottomrule
    \end{tabular}
    }
    \caption{Performance of LLaMA-3-8B after restoring different scales of parameters on XSum~\cite{XSum} (Summarization) and GSM8K~\cite{GSM8K} (Math).}
    \label{tab:more_task}
\end{table}

\paragraph{\textit{Finding 1}}
\textit{The majority of parameter updates introduced by SFT are unnecessary and can significantly degrade model knowledge.}\footnote{More discussion can be found in Appendix~\ref{sec:dis_resundant}.}

Table~\ref{tab:restore} shows that restoring a portion of the model’s parameters to their pre-trained values consistently improves performance, regardless of the fine-tuning dataset. For instance, when fine-tuning with 1,920 samples, restoring 20\% of the parameters enhances the performance of all models. Specifically, the model fine-tuned with \( \mathcal{D}_{train-0}^\mathcal{M} \) achieves a 9.85\% performance gain. Table~\ref{tab:rank} further reveals that over 90\% of the total parameter variation is restored at this point.
Importantly, the benefits of parameter restoration generalize across tasks, as shown in Table~\ref{tab:more_task}. However, the degree of improvement depends on the relevance of the task to the model's knowledge. Notably, performance on the training set also improves, suggesting that \textit{many of the parameter updates introduced by SFT neither help fit the training data nor support generalization, and may impair previously learned knowledge}.
Compared to other strategies, restoring redundant parameter updates is an effective and simple method for enhancing model performance, offering useful insights for designing more efficient fine-tuning approaches.\footnote{A comparison of different strategies is presented in Appendix~\ref{sec:strageties}.}

\paragraph{\textit{Finding 2}}
\textit{Models fine-tuned with larger datasets or lower-mastery data are more adversely affected by unnecessary parameter changes during SFT.}

While SFT consistently introduces unnecessary parameter updates that degrade model performance, the extent of this effect depends on the scale and category of fine-tuning data.  
On one hand, models fine-tuned with larger datasets experience a greater impact. Specifically, models trained with 240 samples generally show performance degradation when more than 20\% of the parameters are restored. In contrast, models fine-tuned with 1,920 samples continue to gain performance improvements even after restoring 40\% of the parameters. This suggests that fine-tuning with 1,920 samples introduces a higher proportion of unnecessary updates. Additionally, the maximum performance gain achieved through parameter restoration is greater for models fine-tuned with 1,920 samples than for those fine-tuned with 240 samples.  
On the other hand, models fine-tuned with low-mastery data are also more affected. Regardless of dataset size, models fine-tuned with \( \mathcal{D}_{train-0}^\mathcal{M} \) consistently allow more parameter restoration while achieving greater performance gains compared to other models. For instance, when using 1,920 samples, the model fine-tuned with \( \mathcal{D}_{train-0}^\mathcal{M} \) can restore 40\% of the parameters and achieve a 10.48\% performance gain, whereas the model fine-tuned with \( \mathcal{D}_{train-4}^\mathcal{M} \) achieves a maximum gain of only 3.44\% after restoring 20\% of the parameters.

\section{Conclusion}
In this paper, we conduct an in-depth analysis of five LLMs across two families on the CBQA task, revealing that both the category and scale of fine-tuning data significantly influence performance in unexpected ways. Through token-level analysis, we find that large changes in token logits correlate with degraded model performance, suggesting that excessive parameter updates can harm model knowledge. At the parameter level, we show that up to 90\% of the updates made during SFT are unnecessary or even detrimental for knowledge enhancement. By selectively restoring these updates, we improve model performance while preserving prior knowledge. Our findings challenge conventional fine-tuning practices and offer practical guidance for developing more efficient methods for LLMs.


\section*{Limitations}
Although we conduct an in-depth analysis of anomalies arising from SFT, our work has certain limitations.  
On one hand, the study does not propose a more efficient fine-tuning strategy based on the findings. This is because the focus is on phenomenological analysis to uncover the underlying mechanisms of SFT on model knowledge. Future work should focus on designing adaptive fine-tuning strategies that minimize unnecessary updates while maximizing performance gains. 
On the other hand, due to resource constraints, the analysis is limited to the LLaMA-2 and LLaMA-3 model series. However, preliminary validation on other model families shows that the conclusions generalize, suggesting broader applicability.

\section*{Acknowledgments}
The authors wish to thank the anonymous reviewers for their helpful comments. This work was partially funded by the Science and Technology Commission of Shanghai Municipality (No.24511103100), National Natural Science Foundation of China (No.62476061,62206057), Shanghai Rising-Star Program (23QA1400200), Natural Science Foundation of Shanghai (23ZR1403500).

\bibliography{custom}

\clearpage
\onecolumn

\appendix

\section{Prompt for SFT}
\label{sec:prompt}

To ensure a fair comparison, we use uniform prompt templates during training.

\begin{verbatim}
{% if messages[0]['from'] == 'system' %}
    {% set system_message = '<<SYS>>\n' + messages[0]['value'] | trim + '\n<</SYS>>\n\n' %}
    {% set messages = messages[1:] %}
{% else %}
    {% set system_message = '' %}
{% endif %}
{% for message in messages %}
    {% if (message['from'] == 'user') != (loop.index0 % 2 == 0) %}
        {{ raise_exception('Conversation roles must alternate user/assistant...') }}
    {% endif %}
    {% if loop.index0 == 0 %}
        {% set content = system_message + message['value'] %}
    {% else %}
        {% set content = message['value'] %}
    {% endif %}
    {% if message['from'] == 'user' %}
        {{ bos_token + '<|question|> ' + content | trim + ' <|answer|>' }}
    {% elif message['from'] == 'assistant' %}
        {{ ' ' + content | trim + ' ' + eos_token }}
    {% endif %}
{% endfor %}
\end{verbatim}

\clearpage

\section{More Details of Experiments}
\label{sec:detail}

\subsection{Details of Models}
\label{sec:detail_model}

To ensure generalizable results, we analyze five LLMs from two different families.

\paragraph{LLaMA-2 Family}
The LLaMA-2 family includes three open-source LLMs developed by Meta. These models are pre-trained on over 2 trillion tokens, equipping them with extensive world knowledge and strong semantic representations. For this study, we select \textbf{LLaMA-2-7B}, \textbf{LLaMA-2-13B}, and \textbf{LLaMA-2-70B}.

\paragraph{LLaMA-3 Family}
The LLaMA-3 family builds upon the LLaMA-2 architecture with significant advancements, such as improved parameter efficiency and task generalization. We analyze \textbf{LLaMA-3-8B} and \textbf{LLaMA-3-70B}.

\subsection{Details of Parameter Restoration}
\label{sec:restore-calcu}

To assess how excessive parameter updates affect model performance, we compare the fine-tuned model with the pre-trained model by ranking parameters according to their relative change.

For each parameter $i$, let $p_i$ denote its value before fine-tuning and $s_i$ its value afterward. The relative change is defined as:

$$
r_i = \frac{|s_i - p_i|}{|p_i|}
$$

We sort all parameters in descending order of $r_i$ to obtain the set $I$.

To measure the concentration of parameter updates, we compute the cumulative sum of $r_i$ for the top percentage of parameters in $I$, divided by the total sum of all $r_i$. For instance, Table~\ref{tab:rank} shows that the top 1\% of parameters contribute 70.59\% of the total relative change.

\clearpage

\section{More Results}
\label{sec:more-results}

In this section, we present additional experimental results that are not included in the main body of the paper due to the limitation of space.

\subsection{Test Results for the LLaMA-2 Family Models}
\label{sec:result-llama2}

\begin{figure*}[h]
    \centering
    \begin{subfigure}{0.48\linewidth}
        \centering
        \includegraphics[width=\linewidth]{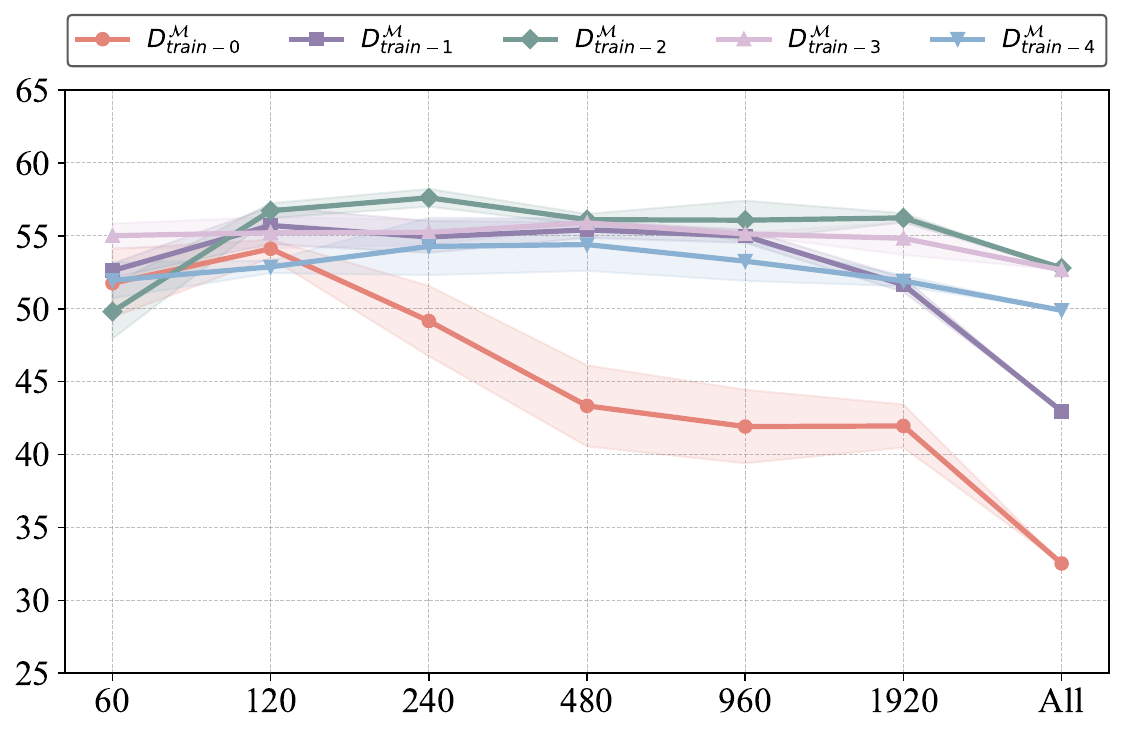}
        \caption{LLaMA-2-7B (In-Domain)}
        \label{fig:result-id-llama-2-7b}
    \end{subfigure}
    \quad
    \begin{subfigure}{0.48\linewidth}
        \centering
        \includegraphics[width=\linewidth]{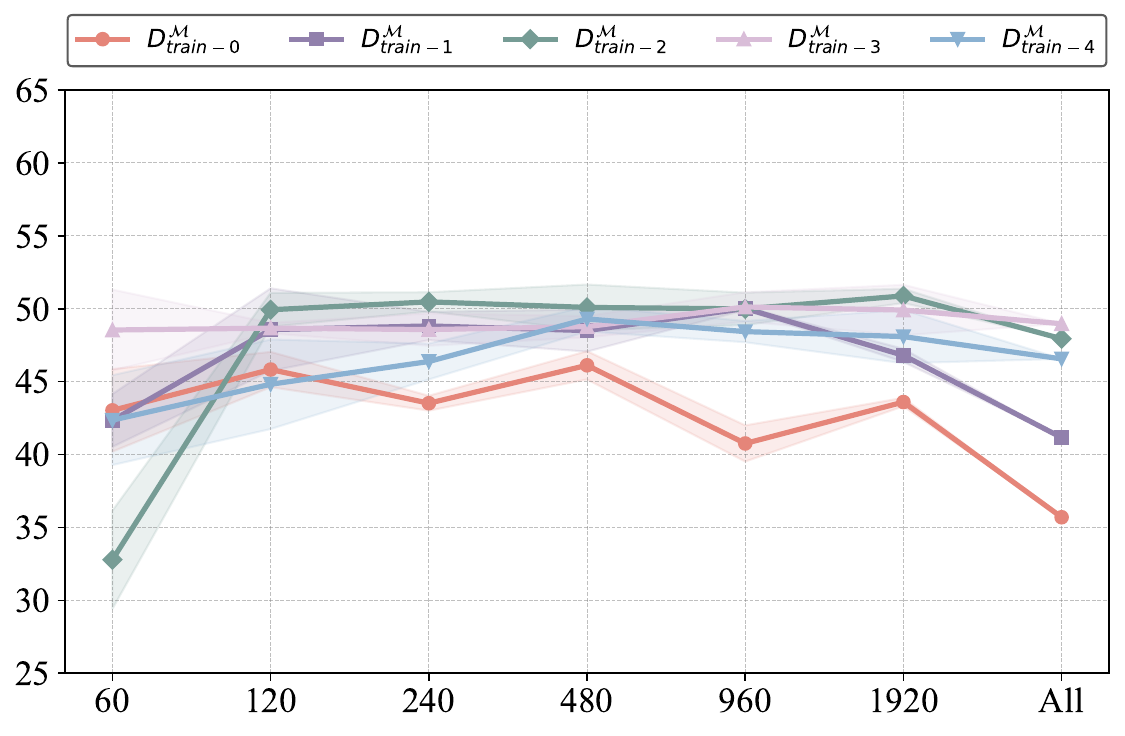}
        \caption{LLaMA-2-7B (Out-of-Domain)}
        \label{fig:result-ood-llama-2-7b}
    \end{subfigure}
    \quad
    \begin{subfigure}{0.48\linewidth}
        \centering
        \includegraphics[width=\linewidth]{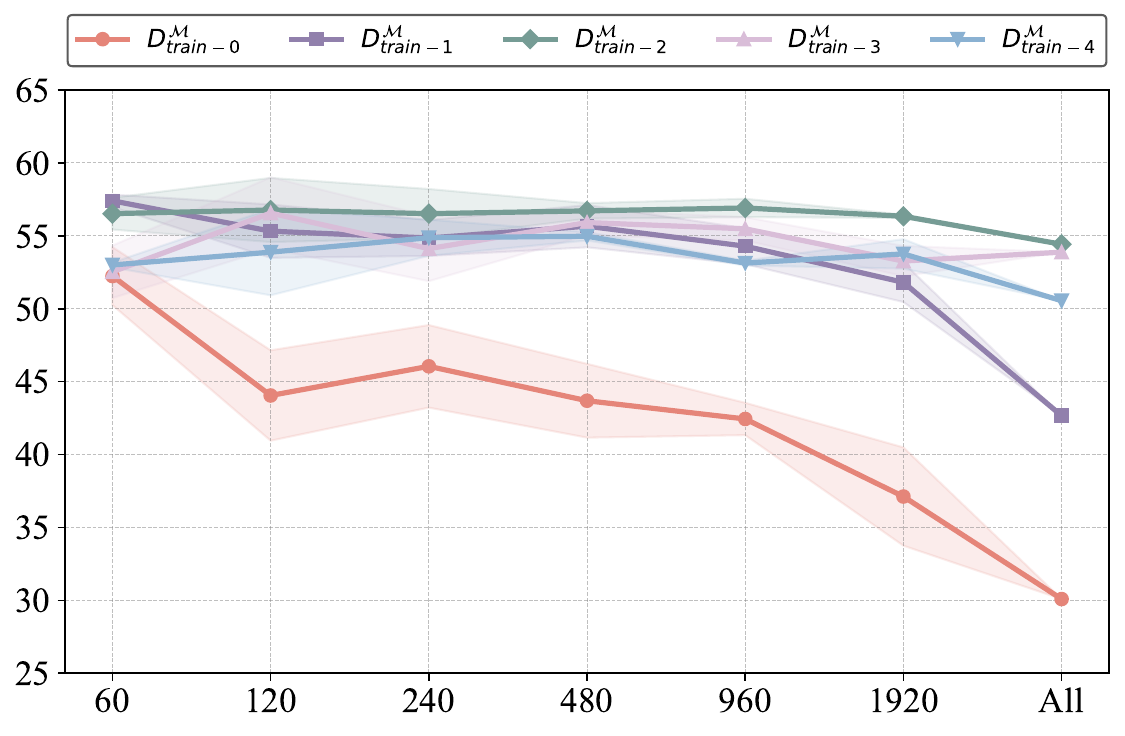}
        \caption{LLaMA-2-13B (In-Domain)}
        \label{fig:result-id-llama-2-13b}
    \end{subfigure}
    \quad
    \begin{subfigure}{0.48\linewidth}
        \centering
        \includegraphics[width=\linewidth]{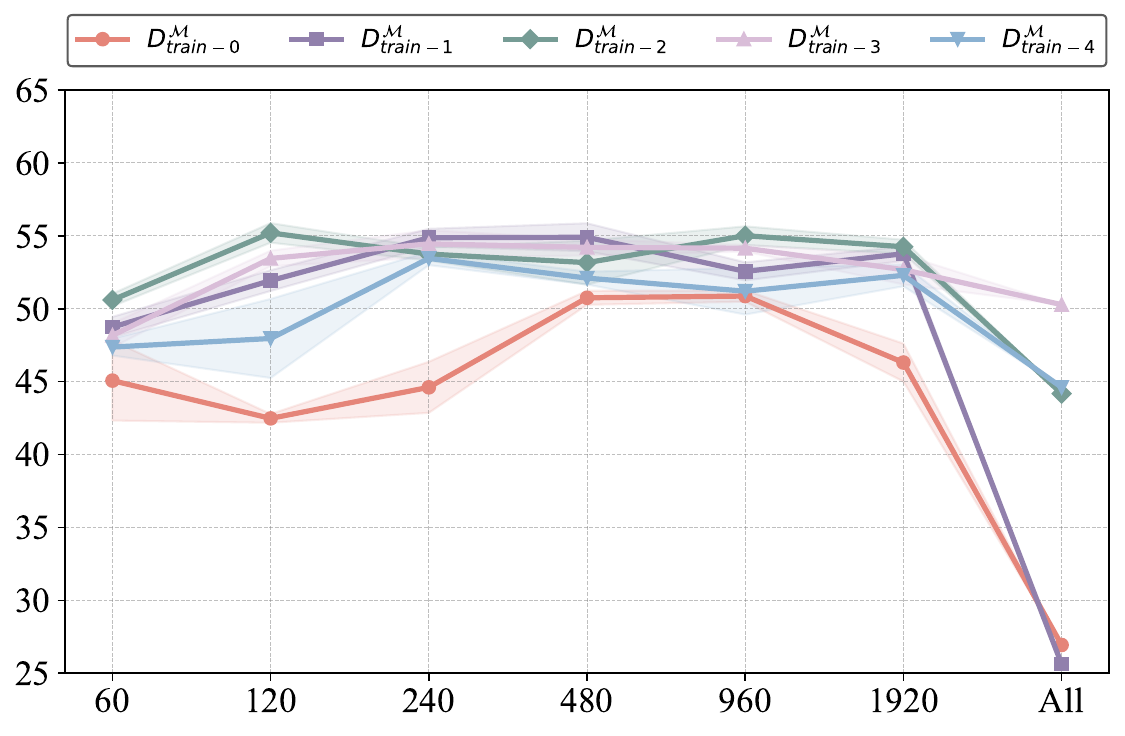}
        \caption{LLaMA-2-13B (Out-of-Domain)}
        \label{fig:result-ood-llama-2-13b}
    \end{subfigure}
    \quad
    \begin{subfigure}{0.48\linewidth}
        \centering
        \includegraphics[width=\linewidth]{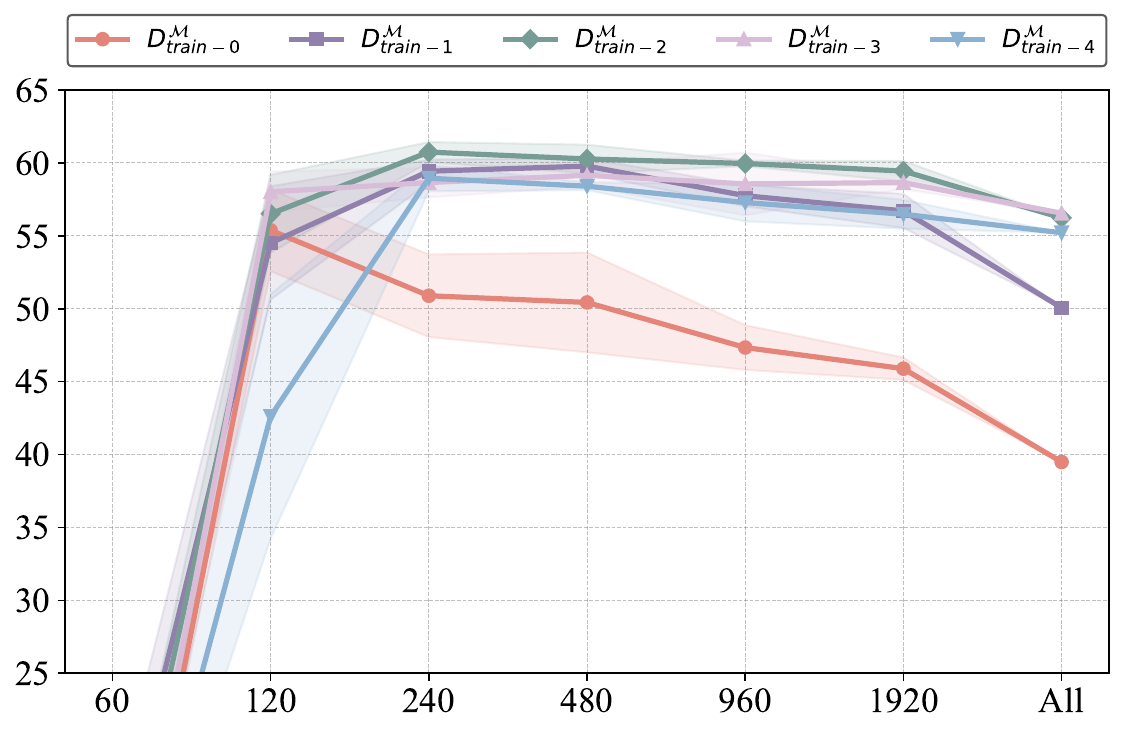}
        \caption{LLaMA-2-70B (In-Domain)}
        \label{fig:result-id-llama-2-70b}
    \end{subfigure}
    \quad
    \begin{subfigure}{0.48\linewidth}
        \centering
        \includegraphics[width=\linewidth]{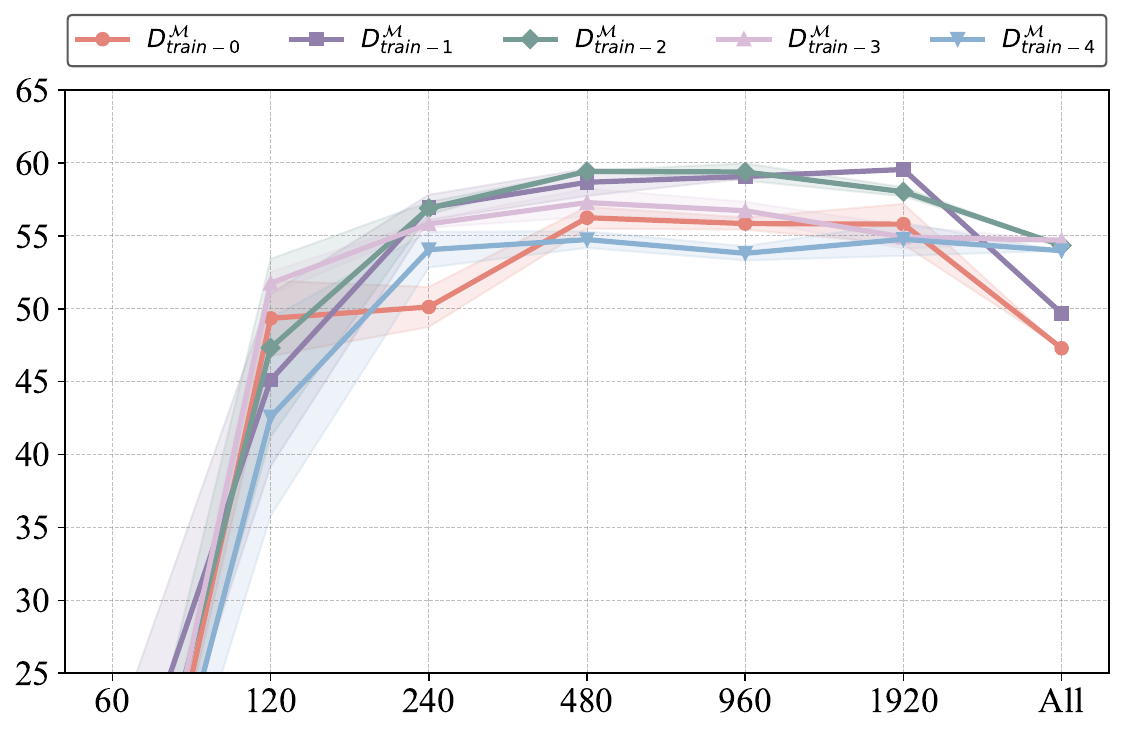}
        \caption{LLaMA-2-70B (Out-of-Domain)}
        \label{fig:result-ood-llama-2-70b}
    \end{subfigure}
    
    \caption{In-domain ($\textbf{Acc}_{test}^\mathcal{M}$) and out-of-domain ($\textbf{Acc}_{testood}^\mathcal{M}$) performance of the LLaMA-3 family models fine-tuned with varying data scales, where `All' indicates the use of the entire dataset listed in Appendix~\ref{sec:detail_distribution}.}
    \label{fig:result-llama-2}
\end{figure*}

We fine-tune five LLMs using datasets with five different levels of mastery. The results for the LLaMA-3 family models are presented in Section~\ref{sec:results}, while the results for the LLaMA-2 family are shown in Figure~\ref{fig:result-llama-2}. Notably, although the peak performance occurs at different data sizes depending on the base model and hyperparameters, the trend of performance degradation beyond a certain point (size) remains consistent.

\clearpage

\subsection{Performance on the Training Set}
\label{sec:result-train}

We compare the performance of different LLMs fin-tuned from LLaMA-3-8B on their respective training sets when restoring different proportions of parameters. The results in Table~\ref{tab:restore_train} show that parameter reduction improves model performance on the training set, further supporting the idea that SFT introduces a significant number of unnecessary or even detrimental parameter updates.

\begin{table}[h]
    \centering
    {
    \begin{tabular}{lccccc}
    \toprule
    \textbf{Restore}  & $\mathbf{\mathcal{D}_{train-0}^\mathcal{M}}$ & $\mathbf{\mathcal{D}_{train-1}^\mathcal{M}}$ & $\mathbf{\mathcal{D}_{train-2}^\mathcal{M}}$ & $\mathbf{\mathcal{D}_{train-3}^\mathcal{M}}$ & $\mathbf{\mathcal{D}_{train-4}^\mathcal{M}}$ \\ \midrule
    \rowcolor{gray!15} \multicolumn{6}{c}{\textit{Number of Training Data: 240}} \\
    0 & 12.08 & 61.25 & 84.58 & 90.00 & 92.92 \\ 
    5\% & \cellcolor{GREEN!5}12.50 & \cellcolor{GREEN!17}62.92 & \cellcolor{GREEN!5}85.00 & \cellcolor{GREEN!8}90.83 & \cellcolor{GREEN!8}93.75 \\
    20\% & \cellcolor{RED!1}11.25 & \cellcolor{GREEN!8}62.08 & \cellcolor{RED!1}83.75 & \cellcolor{GREEN!25}92.5 & \cellcolor{RED!10}82.92 \\
    \midrule
    \rowcolor{gray!15} \multicolumn{6}{c}{\textit{Number of Training Data: 1920}} \\
    0 & 16.56 & 62.81 & 83.44 & 89.48 & 93.39 \\ 
    5\% & \cellcolor{RED!1}15.68 & \cellcolor{GREEN!20}64.74 & \cellcolor{GREEN!21}85.52 & \cellcolor{GREEN!10}90.47 & \cellcolor{GREEN!9}94.22 \\
    20\% & \cellcolor{RED!1}15.16 & \cellcolor{GREEN!22}65.00 & \cellcolor{GREEN!56}89.06 & \cellcolor{GREEN!11}90.57 & \cellcolor{GREEN!17}94.90 \\
    \bottomrule
    \end{tabular}
    }
    \caption{Performance of LLaMA-3-8B on the \textbf{training set} after restoring different scales of parameters across various fine-tuning datasets. Improvements over the non-restored model are highlighted in \colorbox{GREEN}{green}, while performance declines are shown in \colorbox{RED}{red}, with darker shades indicating larger differences.}
    \label{tab:restore_train}
\end{table}

\subsection{Comparison of Results Across Different Strategies}
\label{sec:strageties}
We compare the performance of LLaMA-3-8B trained using four different strategies:  

\begin{itemize}
    \item \textbf{LLaMA-3-8B-Instruct}: A chat-optimized version fine-tuned by Meta, demonstrating strong performance across various benchmarks.  
    \item \textbf{SFT (Mixed)}: Fine-tuning LLaMA-3-8B using a randomly mixed dataset. Results are tested across different data volumes, with the best outcomes reported.  
    \item \textbf{SFT (Divided)}: Fine-tuning LLaMA-3-8B with data divided based on the model’s mastery level. The best results are reported when fine-tuning with 1,920 samples.  
    \item  \textbf{LoRA}: Fine-tuning LLaMA-3-8B using a randomly mixed dataset with LoRA~\cite{LoRA}.
    \item \textbf{Parameter Restore}: Fine-tuning LLaMA-3-8B using the divided dataset, followed by a parameter restoration process. The best results are reported when fine-tuning with 1,920 samples.  
\end{itemize}

The results in Table~\ref{tab:strageties} indicate that data division and parameter restoration strategies significantly enhance model performance, offering valuable insights for optimizing data selection and fine-tuning approaches.

\begin{table}[h]
    \centering
    \resizebox{\linewidth}{!}
    {
    \begin{tabular}{lccccc}
    \toprule
    \textbf{Strategies}   & \textbf{LLaMA-3-8B-Instruct}  & \textbf{SFT (Mixed)} & \textbf{SFT (Divided)} & \textbf{LoRA} & \textbf{Parameter Restoration}  \\ \midrule
    $\textbf{Acc}_{test}^\mathcal{M}$     & 53.83 &  58.67 & 58.80 & 57.82 & \textbf{62.21} \\
    $\textbf{Acc}_{testood}^\mathcal{M}$ & 54.14 & 53.88 & 54.04 & 51.52 & \textbf{57.10} \\ \bottomrule
    \end{tabular}
    }
    \caption{Performance of different LLMs fine-tuned using various strategies. The best results are highlighted in \textbf{bold}.}
    \label{tab:strageties}
\end{table}

\clearpage

\section{Data Distribution of Different LLMs}
\label{sec:detail_distribution}

Since data division is based on the model's mastery of the data, we analyze the data distributions corresponding to different pre-trained LLMs. The results for LLaMA-3-8B are presented in Section~\ref{sec:dataset}, while the distributions for other models are shown in Table~\ref{tab:data-other}.

\begin{table}[h]
    \centering
    \begin{subtable}{0.48\linewidth}
        \centering
        \resizebox{\linewidth}{!}{
        \begin{tabular}{lccccc}
        \toprule
        \toprule
        \rowcolor{gray!15} {$\mathcal{D}_{train}$} & $\mathcal{D}_{train-0}^\mathcal{M}$ & $\mathcal{D}_{train-1}^\mathcal{M}$ & $\mathcal{D}_{train-2}^\mathcal{M}$ & $\mathcal{D}_{train-3}^\mathcal{M}$ & $\mathcal{D}_{train-4}^\mathcal{M}$ \\ \midrule
        \textbf{Number} & 12530 & 26805 & 14961 & 11542 & 10106 \\ \midrule
        
        \rowcolor{gray!15} $\mathcal{D}_{test}$ & $\mathcal{D}_{test-0}^\mathcal{M}$ & $\mathcal{D}_{test-1}^\mathcal{M}$ & $\mathcal{D}_{test-2}^\mathcal{M}$ & $\mathcal{D}_{test-3}^\mathcal{M}$ & $\mathcal{D}_{test-4}^\mathcal{M}$ \\ \midrule
        \textbf{Number} & 1595 & 3374 & 1876 & 1491 & 1219\\ \midrule
    
        \rowcolor{gray!15} $\mathcal{D}_{testood}$ & $\mathcal{D}_{testood-0}^\mathcal{M}$ & $\mathcal{D}_{testood-1}^\mathcal{M}$ & $\mathcal{D}_{testood-2}^\mathcal{M}$ & $\mathcal{D}_{testood-3}^\mathcal{M}$ & $\mathcal{D}_{testood-4}^\mathcal{M}$ \\ \midrule
             \textbf{Number} & 2795 & 4517 & 1704 & 1542 & 1055 \\
             \bottomrule
             \bottomrule
        \end{tabular}
        }
        \caption{LLaMA-3-70B}
    \end{subtable}
    \quad
    \begin{subtable}{0.48\linewidth}
        \centering
        \resizebox{\linewidth}{!}{
        \begin{tabular}{lccccc}
        \toprule
        \toprule
        \rowcolor{gray!15} {$\mathcal{D}_{train}$} & $\mathcal{D}_{train-0}^\mathcal{M}$ & $\mathcal{D}_{train-1}^\mathcal{M}$ & $\mathcal{D}_{train-2}^\mathcal{M}$ & $\mathcal{D}_{train-3}^\mathcal{M}$ & $\mathcal{D}_{train-4}^\mathcal{M}$ \\ \midrule
        \textbf{Number} & 22725 & 30566 & 9336 & 7508 & 5809 \\ \midrule
        
        \rowcolor{gray!15} $\mathcal{D}_{test}$ & $\mathcal{D}_{test-0}^\mathcal{M}$ & $\mathcal{D}_{test-1}^\mathcal{M}$ & $\mathcal{D}_{test-2}^\mathcal{M}$ & $\mathcal{D}_{test-3}^\mathcal{M}$ & $\mathcal{D}_{test-4}^\mathcal{M}$ \\ \midrule
        \textbf{Number} & 2941 & 3805 & 1162 & 958 & 689\\ \midrule
    
        \rowcolor{gray!15} $\mathcal{D}_{testood}$ & $\mathcal{D}_{testood-0}^\mathcal{M}$ & $\mathcal{D}_{testood-1}^\mathcal{M}$ & $\mathcal{D}_{testood-2}^\mathcal{M}$ & $\mathcal{D}_{testood-3}^\mathcal{M}$ & $\mathcal{D}_{testood-4}^\mathcal{M}$ \\ \midrule
             \textbf{Number} & 5201 & 4181 & 1030 & 786 & 415 \\
             \bottomrule
             \bottomrule
        \end{tabular}
        }
        \caption{LLaMA-2-7B}
    \end{subtable}
    \quad
    \begin{subtable}{0.48\linewidth}
        \centering
        \resizebox{\linewidth}{!}{
        \begin{tabular}{lccccc}
        \toprule
        \toprule
        \rowcolor{gray!15} {$\mathcal{D}_{train}$} & $\mathcal{D}_{train-0}^\mathcal{M}$ & $\mathcal{D}_{train-1}^\mathcal{M}$ & $\mathcal{D}_{train-2}^\mathcal{M}$ & $\mathcal{D}_{train-3}^\mathcal{M}$ & $\mathcal{D}_{train-4}^\mathcal{M}$ \\ \midrule
        \textbf{Number} & 20899 & 30562 & 9798 & 7996 & 6689 \\ \midrule
        
        \rowcolor{gray!15} $\mathcal{D}_{test}$ & $\mathcal{D}_{test-0}^\mathcal{M}$ & $\mathcal{D}_{test-1}^\mathcal{M}$ & $\mathcal{D}_{test-2}^\mathcal{M}$ & $\mathcal{D}_{test-3}^\mathcal{M}$ & $\mathcal{D}_{test-4}^\mathcal{M}$ \\ \midrule
        \textbf{Number} & 2675 & 3791 & 1275 & 1006 & 808\\ \midrule
    
        \rowcolor{gray!15} $\mathcal{D}_{testood}$ & $\mathcal{D}_{testood-0}^\mathcal{M}$ & $\mathcal{D}_{testood-1}^\mathcal{M}$ & $\mathcal{D}_{testood-2}^\mathcal{M}$ & $\mathcal{D}_{testood-3}^\mathcal{M}$ & $\mathcal{D}_{testood-4}^\mathcal{M}$ \\ \midrule
             \textbf{Number} & 4671 & 4242 & 1233 & 981 & 486 \\
             \bottomrule
             \bottomrule
        \end{tabular}
        }
        \caption{LLaMA-2-13B}
    \end{subtable}
    \quad
    \begin{subtable}{0.48\linewidth}
        \centering
        \resizebox{\linewidth}{!}{
        \begin{tabular}{lccccc}
        \toprule
        \toprule
        \rowcolor{gray!15} {$\mathcal{D}_{train}$} & $\mathcal{D}_{train-0}^\mathcal{M}$ & $\mathcal{D}_{train-1}^\mathcal{M}$ & $\mathcal{D}_{train-2}^\mathcal{M}$ & $\mathcal{D}_{train-3}^\mathcal{M}$ & $\mathcal{D}_{train-4}^\mathcal{M}$ \\ \midrule
        \textbf{Number} & 15378 & 29468 & 13385 & 9344 & 8369 \\ \midrule
        
        \rowcolor{gray!15} $\mathcal{D}_{test}$ & $\mathcal{D}_{test-0}^\mathcal{M}$ & $\mathcal{D}_{test-1}^\mathcal{M}$ & $\mathcal{D}_{test-2}^\mathcal{M}$ & $\mathcal{D}_{test-3}^\mathcal{M}$ & $\mathcal{D}_{test-4}^\mathcal{M}$ \\ \midrule
        \textbf{Number} & 1956 & 3669 & 1719 & 1199 & 1012\\ \midrule
    
        \rowcolor{gray!15} $\mathcal{D}_{testood}$ & $\mathcal{D}_{testood-0}^\mathcal{M}$ & $\mathcal{D}_{testood-1}^\mathcal{M}$ & $\mathcal{D}_{testood-2}^\mathcal{M}$ & $\mathcal{D}_{testood-3}^\mathcal{M}$ & $\mathcal{D}_{testood-4}^\mathcal{M}$ \\ \midrule
             \textbf{Number} & 3339 & 4537 & 1511 & 1338 & 888 \\
             \bottomrule
             \bottomrule
        \end{tabular}
        }
        \caption{LLaMA-2-70B}
    \end{subtable}
    
    \caption{Data distribution for different LLMs.}
    \label{tab:data-other}
\end{table}

\clearpage

\section{Discussion of Redundant Parameter Updates}
\label{sec:dis_resundant}

\subsection{Distribution of Redundant Parameter Updates}

To investigate why SFT leads to redundant parameter updates, we analyze the distribution of redundant parameters in LLaMA-3-8B. As shown in Table~\ref{tab:dis_layer} and Table~\ref{tab:dis_module}, these parameters are spread across all layers of the model, with a higher concentration in the initial layers (i.e, 0–2), fewer in the final layers (i.e., 30–31), and a more uniform distribution in the middle layers (i.e., 3–29). This pattern may be due to the initial layers primarily encoding semantic information that is already well-learned during pretraining, resulting in greater parameter redundancy. In contrast, the final layers, which focus on output formatting, exhibit less redundancy. Furthermore, we observe that most redundant parameters are concentrated in the FFN layers, suggesting that their high parameter count presents a potential target for optimization. We also acknowledge that the emergence of redundant parameters may be linked to the lottery ticket hypothesis~\cite{LotteryTicket}.

\begin{table}[H]
    \centering
    \resizebox{\linewidth}{!}
    {
    \begin{tabular}{lcccccccccccccccc}
    \toprule
    \toprule
    \rowcolor{gray!15} \textbf{Layer}  &  \textbf{0} & \textbf{1} & \textbf{2} & \textbf{3} & \textbf{4} & \textbf{5} & \textbf{6} & \textbf{7} & \textbf{8} & \textbf{9} & \textbf{10} & \textbf{11} & \textbf{12} & \textbf{13} & \textbf{14} & \textbf{15} \\ \midrule
    \textbf{Percentage} & 3.77 & 3.26 & 3.25 & 3.15 & 3.17 & 3.20 & 3.19 & 3.20 & 3.23 & 3.21 & 3.22 & 3.18 & 3.10 & 3.13 & 3.15 & 3.12 \\ \midrule
    \rowcolor{gray!15} \textbf{Layer}  &  \textbf{16} & \textbf{17} & \textbf{18} & \textbf{19} & \textbf{20} & \textbf{21} & \textbf{22} & \textbf{23} & \textbf{24} & \textbf{25} & \textbf{26} & \textbf{27} & \textbf{28} & \textbf{29} & \textbf{30} & \textbf{31} \\ \midrule
    \textbf{Percentage} & 3.16 & 3.11 & 3.08 & 3.07& 3.10 & 3.06 & 3.06 & 3.03 & 3.01 & 2.99 & 3.00 & 3.03 & 2.98 & 2.95 & 2.99 & 2.84 \\ 
    \bottomrule
    \bottomrule
    \end{tabular}
    }
    \caption{Distribution of redundant parameter updates across layers in LLaMA-3-8B.}
    \label{tab:dis_layer}
\end{table}

\begin{table}[H]
    \centering
    \begin{tabular}{lccccccc}
    \toprule
    \toprule
    \rowcolor{gray!15} \textbf{Module}  &  \textbf{mlp.down} & \textbf{mlp.up} & \textbf{mlp.gate} & \textbf{attn.o} & \textbf{attn.q} & \textbf{attn.v} & \textbf{attn.k}  \\ \midrule
    \textbf{Percentage} & 28.91 & 28.26 & 23.37 & 9.40 & 6.41 & 2.55 & 1.09 \\
    \bottomrule
    \bottomrule
    \end{tabular}
    \caption{Distribution of redundant parameter updates across modules in LLaMA-3-8B.}
    \label{tab:dis_module}
\end{table}

\subsection{Layers for Preserving Model Behavior}

To identify which layers are most critical for preserving model behavior, we perform experiments that selectively restore parameters in different layers and evaluate the resulting performance.
As shown in Table~\ref{tab:restored-layers}, our results reveal that the lower layers (0–3) are most crucial for maintaining model behavior, while the middle layers (4–27) have the least impact. This implies that SFT affects not only the predictive distributions in the upper layers but also substantially modifies the lower-layer parameters, thereby influencing overall model performance.

\begin{table}[H]
\centering
\begin{tabular}{lcccccccc}
\toprule
\textbf{Restored Layers} & \textbf{0-3} & \textbf{4-7} & \textbf{8-11} & \textbf{12-15} & \textbf{16-19} & \textbf{20-23} & \textbf{24-27} & \textbf{28-31} \\
\midrule
$\mathcal{D}_{\text{train-0}}$ & 9.83  & 57.05 & 55.59 & 55.56 & 55.22 & 55.50 & 55.14 & 57.60 \\
$\mathcal{D}_{\text{train-1}}$ & 21.59 & 58.22 & 57.83 & 58.02 & 58.03 & 57.99 & 57.96 & 57.99 \\
$\mathcal{D}_{\text{train-2}}$ & 53.75 & 56.50 & 59.54 & 59.38 & 59.26 & 59.36 & 58.95 & 34.68 \\
$\mathcal{D}_{\text{train-3}}$ & 2.80  & 59.43 & 59.12 & 59.17 & 59.15 & 59.18 & 59.15 & 50.76 \\
$\mathcal{D}_{\text{train-4}}$ & 4.00  & 53.72 & 54.04 & 53.94 & 53.74 & 53.91 & 53.14 & 47.32 \\
\bottomrule
\end{tabular}
\caption{Model performance when restoring different layers under different datasets.}
\label{tab:restored-layers}
\end{table}

\clearpage

\section{Details of Data Processing}
\label{sec:detail_data}

In this section, we provide additional details on data processing.

\subsection{Robust Multi-Template Complementation Mechanism}
\label{sec:complementation}

As described in~\citet{sft-ye}, consider a knowledge fact \( k \) represented as a triple \( (subject,~relation,~object) \), such as \( (Painblanc,~located in,~France) \). Given a sentence \( x = \text{map}(subject,~relation) \) that maps the subject and relation (e.g., \textit{Painblanc is located in}), an LLM \( \mathcal{M} \) is considered to have memorized \( k \) if it can predict \( y = \text{map}(object) \) by mapping the object (e.g., \textit{France}) such that \( y \subseteq \mathcal{M}(x) \).  

Since \( \mathcal{M} \) is a probabilistic model influenced by different mapping templates and sampling probabilities, we design \( N_{\text{map}} = 21 \) different mappings for each knowledge fact \( k \). With the temperature set to 0.7, the model generates \( N_{\text{sample}} = 10 \) outputs for each mapping. The degree to which the LLM memorizes \( k \) is then calculated as:  
\[
    R_k^\mathcal{M} = \frac{\sum_{i=1}^{N_{\text{map}}} \sum_{j=1}^{N_{\text{sample}}} \mathbb{I}(y_i \subseteq \mathcal{M}^j(x_i))}{N_{\text{map}} \times N_{\text{sample}}}
\]
where \( x_i \) and \( y_i \) represent the results from the \( i \)th mapping, \( \mathcal{M}^j \) denotes the \( j \)th sampled output, and \( \mathbb{I}(\cdot) \) is the indicator function.  

This approach effectively utilizes the characteristics of LLMs to evaluate their mastery of different data. However, as entities often have multiple aliases (e.g., \textit{USA} and \textit{United States}), the singular entity labeling in the original dataset may introduce biases. To enhance robustness, a synonym mapping table (Table~\ref{tab:synonym_mapping}) is constructed to expand the set of equivalent entity names, significantly improving result accuracy.
This table is also used in judging the accuracy of LLMs' answers after SFT.

\begin{table}[h]
    \centering
    \resizebox{0.93\linewidth}{!}{ 
    \begin{tabular}{ll}
        \toprule
        \textbf{Object} & \textbf{Synonyms} \\
        \midrule
        United States of America & USA, United States, United States of America \\
        New York City & New York, New York City \\
        University of Michigan & UMich, University of Michigan \\
        South Korea & South Korea, Republic of Korea, Korea \\
        Saint Petersburg & Saint Petersburg, St. Petersburg \\
        Buenos Aires & Baires, Buenos Aires \\
        People's Republic of China & PRC, People's Republic of China, China \\
        Ohio State University & Ohio State University, Ohio State \\
        Bosnia and Herzegovina & Bosnia, Bosnia and Herzegovina, Bosna i Hercegovina \\
        University of Texas at Austin & University of Texas at Austin, University of Texas, UT Austin \\
        University of Cambridge & Cambridge University, Cambridge, University of Cambridge \\
        United States Military Academy & United States Military Academy, West Point \\
        Rio de Janeiro & Rio de, Rio de Janeiro \\
        University of Edinburgh & Edinburgh University, University of Edinburgh \\
        Museo del Prado & Prado Museum, Museo Nacional del Prado, Museo del Prado \\
        Salt Lake City & Salt Lake, Salt Lake City \\
        North Carolina State University & NC State, North Carolina State University \\
        University of Durham & University of Durham, Durham University \\
        Harvard Law School & Harvard University, Harvard Law School \\
        University of Paris (1896-1968) & Universit\'e de Paris, University of Paris, Paris University \\
        Newcastle upon Tyne & Newcastle upon Tyne, Newcastle \\
        University of Oslo & University of Oslo, Oslo University \\
        Hebrew University of Jerusalem & University of Jerusalem, Hebrew University, Hebrew University of Jerusalem \\
        Carnegie Mellon University & Carnegie Mellon University, Carnegie Mellon \\
        University of Oxford & Oxford University, University of Oxford \\
        Autodromo Nazionale Monza & Monza, Autodromo Nazionale Monza \\
        Indiana State House & Indiana State House, Indiana State \\
        Imperial College London & Imperial College, Imperial College London \\
        United Arab Emirates & UAE, United Arab Emirates, The Emirates \\
        \bottomrule
    \end{tabular}}
    \caption{Synonym mapping table for objects in the dataset.}
    \label{tab:synonym_mapping}
\end{table}

\clearpage

\subsection{Topics and Mapping Templates of Data}
We categorize 10 location-related topics as in-domain data and another 14 unrelated topics as out-of-domain data, designing 21 mapping templates for each topic. The corresponding data details of in-domain data are listed from Table~\ref{tab:P17} to Table~\ref{tab:P740}, while the corresponding data details of out-of-domain data are listed from Table~\ref{tab:P112} to Table~\ref{tab:P800}.

\begin{table}[H]
\centering
\begin{tabular}{p{0.95\linewidth}}
\toprule
\textbf{Topic:} P17 \\ \textbf{Question Template: } Which country is \{subject\} located in? \\  \midrule
\textbf{Mapping Templates:} \\ 
\{subject\} is located in\\ 
The location of \{subject\} is in\\ 
\{subject\} finds its place within the borders of\\ 
The \{subject\} is situated in the country,\\ 
If you're seeking the \{subject\}, look no further than the nation of\\ 
The land encompassing the \{subject\} is known as\\ 
\{subject\} can be found in\\ 
\{subject\} has its roots in\\ 
The place \{subject\} calls home is\\ 
\{subject\} is situated in\\ 
The geographical location of \{subject\} is in\\ 
\{subject\} can be discovered in the nation of\\ 
The country where \{subject\} is found is\\ 
\{subject\}'s location is in\\ 
\{subject\} resides in\\ 
The country of \{subject\} is\\ 
\{subject\} belongs to\\ 
\{subject\} exists in\\ 
You can find \{subject\} in\\ 
\{subject\} is a part of\\ 
\{subject\} lies within the borders of \\ \bottomrule
\end{tabular}
\caption{Information and mapping templates for topic P17 (in-domain).}
\label{tab:P17}
\end{table}

\clearpage

\begin{table}[H]
\centering
\begin{tabular}{p{0.95\linewidth}}
\toprule
\textbf{Topic:} P19 \\ \textbf{Question Template: } Where was \{subject\} born? \\  \midrule
\textbf{Mapping Templates:} \\ 
\{subject\} was born in\\ 
The birthplace of \{subject\} was\\ 
It is known that \{subject\} came into the world in\\ 
\{subject\} entered the world in\\ 
\{subject\} was born, and that location is\\ 
\{subject\}'s life began in\\ 
The location of \{subject\}'s birth is\\ 
\{subject\}'s birth occurred in\\ 
The place where \{subject\} was born is\\ 
\{subject\} hailed from\\ 
The answer to where \{subject\} was born lies in\\ 
\{subject\} originated from\\ 
\{subject\} came into this world in\\ 
\{subject\} entered life in\\ 
\{subject\} first drew breath in\\ 
The origin of \{subject\} is in\\ 
\{subject\} hails from\\ 
The place of birth for \{subject\} is\\ 
\{subject\}'s birth took place in\\ 
When it comes to birth, \{subject\} was born in\\ 
If you were to ask where \{subject\} was born, it would be \\ \bottomrule
\end{tabular}
\caption{Information and mapping templates for topic P19 (in-domain).}
\label{tab:P19}
\end{table}

\begin{table}[H]
\centering
\begin{tabular}{p{0.95\linewidth}}
\toprule
\textbf{Topic:} P20 \\ \textbf{Question Template: } Where did \{subject\} die? \\  \midrule
\textbf{Mapping Templates:} \\ 
\{subject\} met their end at\\ 
\{subject\} breathed their last at\\ 
\{subject\}'s life came to a close at\\ 
The place of \{subject\}'s death is\\ 
The location of \{subject\}'s demise is\\ 
The site of \{subject\}'s final rest is\\ 
The place where \{subject\} passed away is\\ 
\{subject\}'s mortal remains are in\\ 
\{subject\} succumbed to death in\\ 
The destination of \{subject\}'s last days was\\ 
The story of \{subject\}'s life concluded in\\ 
\{subject\} bid farewell to the world from within the confines of\\ 
The final resting place of \{subject\} is\\ 
\{subject\} took his final breath in\\ 
\{subject\}'s life journey ended in\\ 
\{subject\} died in\\ 
The place where \{subject\} died is\\ 
\{subject\}'s death occurred in\\ 
\{subject\} took their last breath in\\ 
When it comes to death, \{subject\} died in\\ 
Looking at the end of \{subject\}'s life, they died in \\ \bottomrule
\end{tabular}
\caption{Information and mapping templates for topic P20 (in-domain).}
\label{tab:P20}
\end{table}

\begin{table}[H]
\centering
\begin{tabular}{p{0.95\linewidth}}
\toprule
\textbf{Topic:} P36 \\ \textbf{Question Template: } What is the capital of \{subject\}? \\  \midrule
\textbf{Mapping Templates:} \\ 
The capital of \{subject\} is\\ 
When considering the capital of \{subject\}, it is\\ 
In \{subject\}, the city designated as the capital is\\ 
\{subject\}'s capital city is\\ 
The capital city of \{subject\} is located in\\ 
\{subject\} is governed from\\ 
The seat of government in \{subject\} is\\ 
\{subject\}'s governmental hub is\\ 
The administrative center of \{subject\} is\\ 
The political heart of \{subject\} beats in\\ 
One can find \{subject\}'s seat of power in the city of\\ 
One would find \{subject\}'s governing institutions nestled within the boundaries of\\ 
\{subject\}'s capital is\\ 
The capital of the region \{subject\} is\\ 
\{subject\}'s capital designation goes to\\ 
The main city of \{subject\} is\\ 
\{subject\} has its capital in\\ 
The chief city of \{subject\} is\\ 
Looking at \{subject\}, its capital is\\ 
In terms of capital cities, \{subject\} has\\ 
As the capital of \{subject\}, you'll find \\ \bottomrule
\end{tabular}
\caption{Information and mapping templates for topic P36 (in-domain).}
\label{tab:P36}
\end{table}

\begin{table}[H]
\centering
\begin{tabular}{p{0.95\linewidth}}
\toprule
\textbf{Topic:} P69 \\ \textbf{Question Template: } Where was \{subject\} educated? \\  \midrule
\textbf{Mapping Templates:} \\ 
\{subject\} received education at\\ 
\{subject\} completed the studies at\\ 
\{subject\} was schooled at\\ 
\{subject\} was educated in\\ 
\{subject\} graduated from\\ 
\{subject\} spent the formative years at\\ 
\{subject\}'s alma mater is\\ 
\{subject\} pursued the studies at\\ 
\{subject\} gained the knowledge at\\ 
The academic journey of \{subject\} took place in\\ 
The institution where \{subject\} studied is\\ 
Education for \{subject\} was pursued within the walls of\\ 
The educational institution attended by \{subject\} is\\ 
\{subject\} is an alumnus/alumna of\\ 
The academic background of \{subject\} includes\\ 
The place where \{subject\} was educated is\\ 
\{subject\} attended school in\\ 
The education of \{subject\} took place in\\ 
The place of \{subject\}'s education is\\ 
\{subject\} received their education from\\ 
In terms of education, \{subject\} was schooled in \\ \bottomrule
\end{tabular}
\caption{Information and mapping templates for topic P69 (in-domain).}
\label{tab:P69}
\end{table}

\begin{table}[H]
\centering
\begin{tabular}{p{0.95\linewidth}}
\toprule
\textbf{Topic:} P131 \\ \textbf{Question Template: } Where is \{subject\} located? \\  \midrule
\textbf{Mapping Templates:} \\ 
The location of \{subject\} is where you'll find\\ 
If you look where \{subject\} is, you'll see\\ 
Where \{subject\} resides, there also is\\ 
\{subject\} is located at\\ 
\{subject\} can be found in\\ 
\{subject\} is positioned at\\ 
\{subject\} is stationed at\\ 
\{subject\} is based at\\ 
\{subject\} is headquartered at\\ 
The current location of \{subject\} is\\ 
One would locate \{subject\} in the vicinity of\\ 
Currently, \{subject\} resides or occupies\\ 
\{subject\} is in\\ 
The geographical position of \{subject\} is\\ 
\{subject\} is placed in\\ 
You can find \{subject\} in\\ 
\{subject\} exists in\\ 
\{subject\} lies in\\ 
The location of \{subject\} is\\ 
\{subject\} is situated in\\ 
\{subject\} resides in \\ \bottomrule
\end{tabular}
\caption{Information and mapping templates for topic P131 (in-domain).}
\label{tab:P131}
\end{table}

\begin{table}[H]
\centering
\begin{tabular}{p{0.95\linewidth}}
\toprule
\textbf{Topic:} P159 \\ \textbf{Question Template: } Where is the headquarter of \{subject\}? \\  \midrule
\textbf{Mapping Templates:} \\ 
The headquarter of \{subject\} is located in\\ 
\{subject\} has its headquarter in\\ 
You can find the headquarter of \{subject\} in\\ 
\{subject\}'s central office is situated in\\ 
The main hub of \{subject\} is\\ 
\{subject\} is headquartered in\\ 
The location of \{subject\}'s headquarter is\\ 
\{subject\}'s headquarter can be found at\\ 
The address of \{subject\}'s headquarter is\\ 
\{subject\}'s headquarters are located at\\ 
The central hub of operations for \{subject\} can be found in\\ 
The administrative heart of \{subject\} resides at\\ 
\{subject\}'s head office is located in\\ 
\{subject\} has its main base in\\ 
\{subject\}'s headquarters can be found in\\ 
The headquarters of \{subject\} is located in\\ 
\{subject\}'s headquarters is in\\ 
The main office of \{subject\} is in\\ 
\{subject\}'s headquarter is located in\\ 
The headquarter of \{subject\} is situated in\\ 
When it comes to headquarters, \{subject\}'s is in \\ \bottomrule
\end{tabular}
\caption{Information and mapping templates for topic P159 (in-domain).}
\label{tab:P159}
\end{table}

\begin{table}[H]
\centering
\begin{tabular}{p{0.95\linewidth}}
\toprule
\textbf{Topic:} P276 \\ \textbf{Question Template: } Where is \{subject\} located? \\  \midrule
\textbf{Mapping Templates:} \\ 
\{subject\} can be found at\\ 
The location of \{subject\} is\\ 
\{subject\} is situated at\\ 
\{subject\} has its base in\\ 
\{subject\} is headquartered in\\ 
\{subject\} operates out of\\ 
The place where \{subject\} is located is\\ 
\{subject\} is positioned at\\ 
The site of \{subject\} is\\ 
\{subject\} can be found in the location\\ 
The whereabouts of \{subject\} are at\\ 
\{subject\} is situated in the place called\\ 
\{subject\} is established in\\ 
The coordinates of \{subject\} point to\\ 
The address of \{subject\} leads to\\ 
\{subject\} is located in\\ 
\{subject\} resides in\\ 
You can find \{subject\} in\\ 
When it comes to location, \{subject\} is in\\ 
Looking at where \{subject\} is, it is in\\ 
In terms of location, \{subject\} is situated in \\ \bottomrule
\end{tabular}
\caption{Information and mapping templates for topic P276 (in-domain).}
\label{tab:P276}
\end{table}

\begin{table}[H]
\centering
\begin{tabular}{p{0.95\linewidth}}
\toprule
\textbf{Topic:} P495 \\ \textbf{Question Template: } Which country was \{subject\} created in? \\  \midrule
\textbf{Mapping Templates:} \\ 
\{subject\} was created in\\ 
The creation of \{subject\} took place in\\ 
The origin of \{subject\} is traced back to\\ 
\{subject\} was born in\\ 
\{subject\} originated from\\ 
\{subject\} was founded in\\ 
\{subject\} was created in the country of\\ 
The country of origin for \{subject\} is\\ 
\{subject\} originated in the country of\\ 
The birthplace of \{subject\} is none other than\\ 
\{subject\}'s formation occurred in the borders of\\ 
Historically, \{subject\} emerged in the country known as\\ 
\{subject\} was conceptualized in\\ 
The country credit for the creation of \{subject\} goes to\\ 
The country that witnessed the creation of \{subject\} is\\ 
The country where \{subject\} was created is\\ 
\{subject\} was made in\\ 
\{subject\} came into being in\\ 
If you were to ask where \{subject\} was created, it would be\\ 
Looking at the origin of \{subject\}, it was created in\\ 
In terms of country of origin, \{subject\} was created in \\ \bottomrule
\end{tabular}
\caption{Information and mapping templates for topic P495 (in-domain).}
\label{tab:P495}
\end{table}

\begin{table}[H]
\centering
\begin{tabular}{p{0.95\linewidth}}
\toprule
\textbf{Topic:} P740 \\ \textbf{Question Template: } Where was \{subject\} founded? \\  \midrule
\textbf{Mapping Templates:} \\ 
The founding of \{subject\} took place in\\ 
\{subject\} was originally established in\\ 
\{subject\}'s origin is traced back to\\ 
\{subject\} was founded in\\ 
\{subject\} originated in\\ 
\{subject\} has its roots in\\ 
The founding location of \{subject\} is\\ 
\{subject\} has its origins in\\ 
The birthplace of \{subject\} is\\ 
One can trace \{subject\}'s beginnings to\\ 
\{subject\} came into existence in\\ 
The roots of \{subject\} dig deep into the soil of\\ 
\{subject\} traces its beginnings back to\\ 
The inception of \{subject\} took place in\\ 
\{subject\} was brought into existence in\\ 
The founding place of \{subject\} is\\ 
The origin of \{subject\} is in\\ 
The establishment of \{subject\} occurred in\\ 
If you were to ask where \{subject\} was founded, it would be\\ 
Looking at the origin of \{subject\}, it was founded in\\ 
In terms of its founding location, \{subject\} was established in \\ \bottomrule
\end{tabular}
\caption{Information and mapping templates for topic P740 (in-domain).}
\label{tab:P740}
\end{table}

\begin{table}[H]
\centering
\begin{tabular}{p{0.95\linewidth}}
\toprule
\textbf{Topic:} P112 \\ \textbf{Question Template: } Who founded \{subject\}? \\  \midrule
\textbf{Mapping Templates:} \\ 
The founder of \{subject\} is\\ 
\{subject\} was founded by\\ 
The establishment of \{subject\} was initiated by\\ 
\{subject\} owes its existence to\\ 
\{subject\} was brought into being by\\ 
\{subject\} is a brainchild of\\ 
\{subject\} was established by\\ 
\{subject\} has its roots in\\ 
The person who founded \{subject\} is\\ 
The visionary behind \{subject\}'s establishment is\\ 
The inception of \{subject\} can be traced back to\\ 
The idea and realization of \{subject\} were the brainchild of\\ 
\{subject\} was brought into existence by\\ 
\{subject\}'s founder is known to be\\ 
\{subject\} owes its inception to\\ 
\{subject\} was created by\\ 
The creation of \{subject\} is attributed to\\ 
\{subject\} was started by\\ 
\{subject\} originated with\\ 
\{subject\}'s origins lie with\\ 
\{subject\} can trace its roots back to \\ \bottomrule
\end{tabular}
\caption{Information and mapping templates for topic P112 (out-of-domain).}
\label{tab:P112}
\end{table}

\begin{table}[H]
\centering
\begin{tabular}{p{0.95\linewidth}}
\toprule
\textbf{Topic:} P127 \\ \textbf{Question Template: } Who owns \{subject\}? \\  \midrule
\textbf{Mapping Templates:} \\ 
The owner of \{subject\} is\\ 
\{subject\} is owned by\\ 
Ownership of \{subject\} belongs to\\ 
\{subject\} belongs to\\ 
\{subject\} is in the possession of\\ 
\{subject\} is a property of\\ 
\{subject\} is possessed by\\ 
\{subject\} is under the ownership of\\ 
\{subject\} is held by\\ 
The proprietor of \{subject\} is none other than\\ 
Responsibility for \{subject\} falls under the jurisdiction of\\ 
The property known as \{subject\} is under the stewardship of\\ 
The rights to \{subject\} are held by\\ 
The individual who owns \{subject\} is\\ 
The rightful owner of \{subject\} is identified as\\ 
Ownership of \{subject\} is held by\\ 
The possession of \{subject\} is with\\ 
The entity owning \{subject\} is\\ 
\{subject\}'s owner is\\ 
\{subject\} is in the hands of\\ 
If you're looking for the owner of \{subject\}, it's \\ \bottomrule
\end{tabular}
\caption{Information and mapping templates for topic P127 (out-of-domain).}
\label{tab:P127}
\end{table}

\begin{table}[H]
\centering
\begin{tabular}{p{0.95\linewidth}}
\toprule
\textbf{Topic:} P170 \\ \textbf{Question Template: } Who was \{subject\} created by? \\  \midrule
\textbf{Mapping Templates:} \\ 
\{subject\} was created by\\ 
The creator of \{subject\} was\\ 
The person who created \{subject\} is known as\\ 
\{subject\} was founded by\\ 
\{subject\} owes its creation to\\ 
\{subject\} was developed by\\ 
\{subject\}'s creator is\\ 
\{subject\} was the creation of\\ 
The person behind \{subject\} is\\ 
\{subject\} was brought into existence by\\ 
The originator of \{subject\} is\\ 
The creative force behind \{subject\} is attributed to\\ 
\{subject\} came into existence thanks to\\ 
\{subject\} was brought to life by\\ 
\{subject\} was conceptualized by\\ 
The creation of \{subject\} is attributed to\\ 
The entity responsible for creating \{subject\} is\\ 
\{subject\} was made by\\ 
\{subject\}'s creation is attributed to\\ 
When it comes to creation, \{subject\} was created by\\ 
Looking at the creation of \{subject\}, it was done by \\ \bottomrule
\end{tabular}
\caption{Information and mapping templates for topic P170 (out-of-domain).}
\label{tab:P170}
\end{table}

\begin{table}[H]
\centering
\begin{tabular}{p{0.95\linewidth}}
\toprule
\textbf{Topic:} P175 \\ \textbf{Question Template: } Who performed \{subject\}? \\  \midrule
\textbf{Mapping Templates:} \\ 
The performer of \{subject\} was\\ 
\{subject\} was performed by\\ 
The one responsible for performing \{subject\} was\\ 
\{subject\} was brought to life by\\ 
\{subject\} was presented by\\ 
\{subject\} was executed by\\ 
The artist behind \{subject\} is\\ 
The talent behind \{subject\} is\\ 
The one who performed \{subject\} was\\ 
The one who executed \{subject\} skillfully was\\ 
The artist responsible for \{subject\}'s interpretation was\\ 
The responsibility of performing \{subject\} fell upon\\ 
\{subject\} was enacted by\\ 
The act of \{subject\} was performed by\\ 
\{subject\} was executed on stage by\\ 
The execution of \{subject\} was done by\\ 
\{subject\} was carried out by\\ 
The realization of \{subject\} was by\\ 
\{subject\} had its performance by\\ 
The performance of \{subject\} was done by\\ 
Looking at the performance of \{subject\}, it was done by \\ \bottomrule
\end{tabular}
\caption{Information and mapping templates for topic P175 (out-of-domain).}
\label{tab:P175}
\end{table}

\begin{table}[H]
\centering
\begin{tabular}{p{0.95\linewidth}}
\toprule
\textbf{Topic:} P176 \\ \textbf{Question Template: } Which company is \{subject\} produced by? \\  \midrule
\textbf{Mapping Templates:} \\ 
\{subject\} is produced by\\ 
The producer of \{subject\} is\\ 
The production company behind \{subject\} is\\ 
\{subject\} is created by\\ 
\{subject\} is assembled by\\ 
\{subject\} comes from\\ 
\{subject\} is manufactured by\\ 
The company responsible for \{subject\} is\\ 
\{subject\} is a product of\\ 
The production of \{subject\} falls under the umbrella of\\ 
\{subject\} comes from the production house of\\ 
The production of \{subject\} is handled by none other than\\ 
The company behind the production of \{subject\} is\\ 
The company that crafts \{subject\} is\\ 
Every unit of \{subject\} bears the production mark of\\ 
\{subject\} comes from the company\\ 
The production of \{subject\} is handled by\\ 
The company responsible for producing \{subject\} is\\ 
The company that produces \{subject\} is\\ 
When it comes to production, \{subject\} is produced by\\ 
Looking at the production of \{subject\}, it is done by \\ \bottomrule
\end{tabular}
\caption{Information and mapping templates for topic P176 (out-of-domain).}
\label{tab:P176}
\end{table}

\begin{table}[H]
\centering
\begin{tabular}{p{0.95\linewidth}}
\toprule
\textbf{Topic:} P26 \\ \textbf{Question Template: } Who is \{subject\} married to? \\  \midrule
\textbf{Mapping Templates:} \\ 
\{subject\}'s spouse is\\ 
\{subject\} has been married to\\ 
\{subject\} is in a marital union with\\ 
The person \{subject\} is married to is\\ 
\{subject\}'s partner in marriage is\\ 
\{subject\}'s better half is\\ 
\{subject\} is wed to\\ 
\{subject\} exchanged vows with\\ 
\{subject\} shares a life with\\ 
\{subject\} shares a marital bond with\\ 
Their love story culminated in a wedding, uniting \{subject\} and\\ 
The answer to \{subject\}'s nuptials lies in the presence of\\ 
\{subject\} is married to\\ 
\{subject\} has tied the knot with\\ 
\{subject\} shares a matrimonial life with\\ 
The spouse of \{subject\} is\\ 
\{subject\} is wedded to\\ 
In marriage, \{subject\} is united with\\ 
The one \{subject\} is married to is\\ 
\{subject\}'s husband/wife is\\ 
When it comes to marriage, \{subject\} is married to \\ \bottomrule
\end{tabular}
\caption{Information and mapping templates for topic P26 (out-of-domain).}
\label{tab:P26}
\end{table}

\begin{table}[H]
\centering
\begin{tabular}{p{0.95\linewidth}}
\toprule
\textbf{Topic:} P40 \\ \textbf{Question Template: } Who is \{subject\}'s child? \\  \midrule
\textbf{Mapping Templates:} \\ 
The child of \{subject\} is known to be\\ 
Belonging to \{subject\} as a child is\\ 
As a child to \{subject\}, there is\\ 
\{subject\}'s child is\\ 
\{subject\} is the parent of\\ 
\{subject\}'s offspring is\\ 
\{subject\}'s youngster is\\ 
\{subject\}'s family includes\\ 
\{subject\}'s lineage includes\\ 
\{subject\} has a child named\\ 
The offspring of \{subject\} is identified as\\ 
The child of \{subject\} is recognized as\\ 
The offspring of \{subject\} includes\\ 
\{subject\} is the biological parent of\\ 
\{subject\} is the father/mother to\\ 
The child of \{subject\} is\\ 
The progeny of \{subject\} is\\ 
The next generation of \{subject\} includes\\ 
If you were to ask who \{subject\}'s child is, it's\\ 
Looking at \{subject\}'s offspring, it's\\ 
In terms of children, \{subject\} has \\ \bottomrule
\end{tabular}
\caption{Information and mapping templates for topic P40 (out-of-domain).}
\label{tab:P40}
\end{table}

\begin{table}[H]
\centering
\begin{tabular}{p{0.95\linewidth}}
\toprule
\textbf{Topic:} P413 \\ \textbf{Question Template: } What position does \{subject\} play? \\  \midrule
\textbf{Mapping Templates:} \\ 
\{subject\} plays\\ 
The position of \{subject\} is\\ 
In the team, \{subject\} holds the position of\\ 
\{subject\} plays the position of\\ 
\{subject\}'s role is\\ 
\{subject\} is a\\ 
The position played by \{subject\} is\\ 
\{subject\} holds the position of\\ 
\{subject\} is a player in the position of\\ 
In the game, \{subject\} assumes the role of\\ 
\{subject\} is known for the position as\\ 
When playing, \{subject\} takes up the position of\\ 
The role \{subject\} takes on is\\ 
\{subject\} is assigned to the position\\ 
The position that \{subject\} occupies is\\ 
\{subject\} occupies the position of\\ 
\{subject\} fulfills the role of\\ 
\{subject\} is positioned as a\\ 
The position that \{subject\} plays is\\ 
\{subject\}'s position is\\ 
If you were to ask what position \{subject\} plays, it's \\ \bottomrule
\end{tabular}
\caption{Information and mapping templates for topic P413 (out-of-domain).}
\label{tab:P413}
\end{table}

\begin{table}[H]
\centering
\begin{tabular}{p{0.95\linewidth}}
\toprule
\textbf{Topic:} P50 \\ \textbf{Question Template: } Who is the author of \{subject\}? \\  \midrule
\textbf{Mapping Templates:} \\ 
\{subject\} was authored by\\ 
The writer of \{subject\} is\\ 
The person who authored \{subject\} is\\ 
The author of \{subject\} is\\ 
\{subject\} was written by\\ 
\{subject\} is a work by\\ 
The creator of \{subject\} is\\ 
The person responsible for \{subject\} is\\ 
\{subject\} owes its existence to\\ 
The creative mind behind \{subject\} is none other than\\ 
\{subject\} was penned by the talented writer,\\ 
The work known as \{subject\} was brought to life by the author,\\ 
\{subject\} is a work authored by\\ 
The penname associated with \{subject\} is\\ 
The words of \{subject\} were put together by\\ 
The person who wrote \{subject\} is\\ 
\{subject\} was created by\\ 
\{subject\} was drafted by\\ 
If you were to ask who authored \{subject\}, it was\\ 
Looking at the authorship of \{subject\}, it was written by\\ 
\{subject\} is a creation of \\ \bottomrule
\end{tabular}
\caption{Information and mapping templates for topic P50 (out-of-domain).}
\label{tab:P50}
\end{table}

\begin{table}[H]
\centering
\begin{tabular}{p{0.95\linewidth}}
\toprule
\textbf{Topic:} P136 \\ \textbf{Question Template: } What type of music does \{subject\} play? \\  \midrule
\textbf{Mapping Templates:} \\ 
The music played by \{subject\} is\\ 
When \{subject\} plays music, it is\\ 
The musical style of \{subject\} can be categorized as\\ 
\{subject\}'s sound is characterized as\\ 
\{subject\}'s musical talent lies in\\ 
\{subject\} has a knack for\\ 
\{subject\}'s genre of music is\\ 
\{subject\} is known for playing\\ 
\{subject\}'s music style is\\ 
The genre that \{subject\} excels in is\\ 
When it comes to music, \{subject\} is known for their proficiency in\\ 
The tunes produced by \{subject\} belong to the category of\\ 
\{subject\}'s music falls under the category of\\ 
\{subject\} has a musical style that is categorized as\\ 
The music played by \{subject\} can be described as\\ 
The type of music \{subject\} plays is\\ 
The genre of music \{subject\} plays is\\ 
The style of music \{subject\} plays is\\ 
\{subject\} plays the music type of\\ 
Musically, \{subject\} is known to play\\ 
In terms of musical style, \{subject\} plays \\ \bottomrule
\end{tabular}
\caption{Information and mapping templates for topic P136 (out-of-domain).}
\label{tab:P136}
\end{table}

\begin{table}[H]
\centering
\begin{tabular}{p{0.95\linewidth}}
\toprule
\textbf{Topic:} P106 \\ \textbf{Question Template: } What kind of work does \{subject\} do? \\  \midrule
\textbf{Mapping Templates:} \\ 
\{subject\} is employed in\\ 
\{subject\} earns a living by working as\\ 
\{subject\}'s occupation is\\ 
\{subject\} is engaged in\\ 
\{subject\}'s profession is\\ 
\{subject\} works as a\\ 
\{subject\} makes a living as\\ 
\{subject\} has a career in\\ 
\{subject\} is involved in\\ 
\{subject\} engages in the occupation of\\ 
The work that \{subject\} undertakes is classified as\\ 
The focus of \{subject\}'s employment lies in\\ 
The type of work \{subject\} engages in is\\ 
The work performed by \{subject\} falls under\\ 
The work done by \{subject\} falls under the category of\\ 
The kind of work \{subject\} does is\\ 
\{subject\} operates in the field of\\ 
The work \{subject\} performs is\\ 
When it comes to work, \{subject\} does\\ 
\{subject\} works in the field of\\ 
The work done by \{subject\} is \\ \bottomrule
\end{tabular}
\caption{Information and mapping templates for topic P106 (out-of-domain).}
\label{tab:P106}
\end{table}

\begin{table}[H]
\centering
\begin{tabular}{p{0.95\linewidth}}
\toprule
\textbf{Topic:} P264 \\ \textbf{Question Template: } What music label is \{subject\} represented by? \\  \midrule
\textbf{Mapping Templates:} \\ 
\{subject\} is represented by\\ 
The music label representing \{subject\} is\\ 
Regarding representation, \{subject\} is under\\ 
\{subject\} has a record deal with\\ 
\{subject\} has a musical partnership with\\ 
\{subject\}'s music is released by\\ 
\{subject\} is signed to\\ 
\{subject\} is affiliated with\\ 
\{subject\} has a contract with\\ 
\{subject\} is represented by the music label\\ 
The talented \{subject\} is associated with the music label\\ 
\{subject\}'s discography is managed by the renowned label\\ 
\{subject\} is under contract with the music label\\ 
\{subject\} is affiliated with the music label\\ 
The music label backing \{subject\} is\\ 
\{subject\} is signed with the music label\\ 
\{subject\} works with the music label\\ 
\{subject\} is under the music label\\ 
The music label that represents \{subject\} is\\ 
\{subject\} has representation from\\ 
If you were to ask what music label represents \{subject\}, it is \\ \bottomrule
\end{tabular}
\caption{Information and mapping templates for topic P264 (out-of-domain).}
\label{tab:P264}
\end{table}

\begin{table}[H]
\centering
\begin{tabular}{p{0.95\linewidth}}
\toprule
\textbf{Topic:} P407 \\ \textbf{Question Template: } Which language was \{subject\} written in? \\  \midrule
\textbf{Mapping Templates:} \\ 
\{subject\} was originally written in\\ 
The language used for writing \{subject\} was\\ 
The original text of \{subject\} appeared in\\ 
\{subject\} was penned in\\ 
The language of \{subject\} is\\ 
\{subject\} was composed in\\ 
\{subject\} was created in\\ 
\{subject\} is written in the language of\\ 
The writing language of \{subject\} is\\ 
\{subject\} was composed in the language known as\\ 
The linguistic medium of \{subject\} is\\ 
The choice of language for \{subject\} is\\ 
\{subject\} was written in the language of\\ 
The language used to write \{subject\} is\\ 
The original language of \{subject\} is\\ 
The writing of \{subject\} is in\\ 
\{subject\} is composed in\\ 
The text of \{subject\} is in\\ 
\{subject\} was written in\\ 
If you were to ask what language \{subject\} was written in, it's\\ 
Looking at the language of \{subject\}, it's \\ \bottomrule
\end{tabular}
\caption{Information and mapping templates for topic P407 (out-of-domain).}
\label{tab:P407}
\end{table}

\begin{table}[H]
\centering
\begin{tabular}{p{0.95\linewidth}}
\toprule
\textbf{Topic:} P800 \\ \textbf{Question Template: } What is \{subject\} famous for? \\  \midrule
\textbf{Mapping Templates:} \\ 
\{subject\} is famous for\\ 
The fame of \{subject\} is due to\\ 
People recognize \{subject\} for\\ 
\{subject\} is renowned for\\ 
\{subject\}'s claim to fame is\\ 
\{subject\} is celebrated for\\ 
\{subject\} is known for\\ 
\{subject\} is distinguished by\\ 
\{subject\} is admired for\\ 
Fame comes to \{subject\} due to\\ 
Among its achievements, \{subject\} is celebrated for\\ 
\{subject\}'s popularity largely stems from\\ 
\{subject\}'s notable recognition comes from\\ 
\{subject\} is celebrated widely due to\\ 
The fame of \{subject\} is attributed to\\ 
The reason \{subject\} is famous is\\ 
\{subject\} is well-known for\\ 
\{subject\} gained fame for\\ 
If you were to ask what \{subject\} is famous for, it's\\ 
Looking at what made \{subject\} famous, it's\\ 
In terms of fame, \{subject\} is associated with \\ \bottomrule
\end{tabular}
\caption{Information and mapping templates for topic P800 (out-of-domain).}
\label{tab:P800}
\end{table}



\end{document}